\documentclass[conference]{IEEEtran}

\usepackage{times}  
\def\baselinestretch{0.99}

\usepackage{amssymb}
\usepackage{graphicx}
\usepackage{url}
\usepackage{xcolor}
\usepackage{verbatim}
\usepackage[parfill]{parskip}
\usepackage{algpseudocode,algorithm}
\usepackage[bookmarks=true,pdfpagelabels=false,plainpages=true]{hyperref}
\hypersetup{
    unicode=false,          
    pdftoolbar=true,        
    pdfmenubar=true,        
    pdffitwindow=false,     
    pdfstartview={FitH},    
    pdfnewwindow=true,      
    colorlinks=true,       
    filecolor=magenta,      
    linkcolor={red!50!black},
    citecolor={blue!50!black},
    urlcolor={blue!80!black},
    bookmarksopen=true,
    bookmarksnumbered=true
}
\usepackage[all]{hypcap}
\usepackage{etoolbox}
\usepackage{multirow}
\usepackage{rotating}
\usepackage{subcaption}

\makeatletter
\patchcmd{\@makecaption}
  {\scshape}
  {}
  {}
  {}

\patchcmd{\@makecaption}
  {\\}
  {.\ }
  {}
  {}

\newcommand\thefontsize[1]{{#1 The current font size is: \f@size pt\par}}
\makeatother
\def\tablename{Table}

\algnewcommand\algorithmicinput{\textbf{Input:}}
\algnewcommand\Input{\item[\algorithmicinput]}
\algnewcommand\algorithmicoutput{\textbf{Output:}}
\algnewcommand\Output{\item[\algorithmicoutput]}

\renewcommand\textfloatsep{3pt}
\renewcommand\floatsep{3pt}
\renewcommand\intextsep{3pt}
\renewcommand\dbltextfloatsep{3pt}
\renewcommand\dblfloatsep{3pt}

\let\labelindent\relax
\usepackage{enumitem}
\setlist{nosep}

\usepackage{ntheorem}

\makeatletter
\renewtheoremstyle{plain}
  {\item{\theorem@headerfont ##1\ ##2\theorem@separator}~}
  {\item{\theorem@headerfont ##1\ ##2\ (##3)\theorem@separator}~}
\makeatother

{\theoremheaderfont{\upshape\bfseries}
 \theorembodyfont{\normalfont\slshape}
 }

\def\IEEEbibitemsep{-3pt plus -1pt}

\begin{document}

\title{Interpretation of Semantic Tweet Representations}

\vspace{-5mm}

\author{
    \IEEEauthorblockN{Ganesh J\IEEEauthorrefmark{1}, Manish Gupta\IEEEauthorrefmark{1}\IEEEauthorrefmark{2}, Vasudeva Varma\IEEEauthorrefmark{1}}
    \IEEEauthorblockA{\IEEEauthorrefmark{1}IIIT Hyderabad, India. ganesh.j@research.iiit.ac.in; \{manish.gupta,vv\}@iiit.ac.in}
    \IEEEauthorblockA{\IEEEauthorrefmark{2}Microsoft, Hyderabad, India. gmanish@microsoft.com}
}


\maketitle

\begin{abstract}
Research in analysis of microblogging platforms is experiencing a renewed surge with a large number of works applying representation learning models for applications like
sentiment analysis, semantic textual similarity computation, hashtag prediction, etc. Although the performance of the representation learning models has been better than the traditional baselines for such tasks, little is known about the elementary properties of a tweet encoded within these representations, or why particular representations work better for certain tasks. 
Our work presented here constitutes the first step in opening the black-box of vector embeddings for tweets.\\
Traditional feature engineering methods for high-level applications have exploited various elementary properties of tweets. We believe that a tweet representation is effective for an application because it meticulously encodes the application-specific elementary properties of tweets. To understand the elementary properties encoded in a tweet representation, we evaluate the representations on the accuracy to which they can model each of those properties such as tweet length, presence of particular words, hashtags, mentions, capitalization, etc.\\ 
Our systematic extensive study of nine supervised and four unsupervised tweet representations against most popular eight textual and five social elementary properties reveal that Bi-directional LSTMs (BLSTMs) and Skip-Thought Vectors (STV) best encode the textual and social properties of tweets respectively. 
FastText is the best model for low resource settings, providing very little degradation with reduction in embedding size.  Finally, we draw interesting insights by correlating the model performance obtained for elementary property prediction tasks with the high-level downstream applications.
\end{abstract}

\section{Introduction} 
\label{sec:intro}
\setlength{\tabcolsep}{2pt}
\begin{table*}
\centering
\scriptsize
\caption{Survey of Unsupervised Representation Learning Models}
\begin{tabular}{|p{2cm}|p{3cm}|p{6cm}|p{6cm}|}
\hline
\textbf{Model} & \textbf{Architecture} & \textbf{Core Idea} & \textbf{Applications Considered in the Paper} \\ \hline
DSSM~\cite{huang13_cikm} & Deep Feed Forward network & Learn a common mapping for query and document & Document ranking \\
\hline
 CDSSM~\cite{shen14_cikm} & Deep Feed Forward Convolutional network & Learn a common mapping for query and document & Document ranking   \\ \hline
 Paragraph2Vec~\cite{le14_icml} & Word2Vec network & Learn document embedding which are good in predicting the words within it & Sentiment analysis, Document retrieval  \\ \hline
 SSWE~\cite{tang16_tkde} & Simple Feed Forward network & Learn sentiment specific word embeddings using distant supervision (emoticons) & Sentiment analysis  \\ \hline
 Skip-Thought vectors~\cite{kiros15_nips} & Gated Recurrent Unit Encoder-Decoder network & Learn sentence embedding which are good in predicting the surrounding sentences (sentential context) & Semantic relatedness, Paraphrase detection, Image-sentence ranking, Sentence classification including sentiment analysis   \\ \hline
 SDAE~\cite{hill16_naacl} & LSTM~\cite{graves13_icassp} Encoder-Decoder network & Predict the source sentence given the corrupted version of the source sentence & Semantic relatedness, Sentence classification tasks used in Skip thought vectors  \\ \hline
 FastSent~\cite{hill16_naacl} & Word2Vec network & Learn sentence embedding which are good in predicting the surrounding sentences (sentential context) & Semantic relatedness, Sentence classification tasks used in Skip thought vectors  \\ \hline
 Siamese CBOW~\cite{kenter16_acl} & Siamese network & Learn sentence embedding which are good in predicting the surrounding sentences (sentential context) & Semantic relatedness  \\ \hline
 Tweet2Vec~\cite{dhingra16_acl} & Bi-GRU Encoder network & Learn tweet embedding directly from characters using hashtags for supervision & Hashtag prediction  \\\hline
\end{tabular}
\label{tab:relatedUnSup}
\end{table*}

Research on Twitter has focused on various kinds of business applications such as opinion mining, semantic textual similarity, user profiling, hashtag identification, microblog retrieval, etc. Central to the performance of these applications~\cite{bengio13_tpami} is the question of tweet representation: How to best capture the essential meaning of a tweet in a machine-understandable format (or ``representation'')? Challenges like short length, informal words, misspellings and unusual grammar make it difficult to obtain a good representation to capture these text aspects. Further, tweets also have social network-oriented properties, and hence a good representation should also capture social aspects. Traditionally, tweets have been modeled using Bag-Of-Words (BOW)~\cite{wang16_acl} and Latent Dirichlet Allocation (LDA)~\cite{wang16_acl}.

Recently there has been a paradigm shift in machine learning towards using distributed representations for words~\cite{mikolov13_nips} and sentences~\cite{le14_icml,hill16_naacl,tai15_acl}. Though these representations are hard to interpret, they have the following advantages: (1) in practice, they are highly effective across multiple applications, and (2) they reduce the dependence on domain level experts. 

Researchers in Twitter analytics have found these representation learning models to be very effective for several critical tasks such as sentiment analysis~\cite{tang16_tkde,giachanou16_csur}, semantic textual similarity computation~\cite{kenter16_acl}, hashtag identification~\cite{dhingra16_acl}, etc. However, little is known about the elementary tweet properties encoded by the representations generated from these models, knowing which will allow us to make generalizable conclusions. Our work presented here constitutes the first step in opening the black-box of vector embeddings for tweets.

Essentially we ask the following question: ``what are the core properties encoded in the given tweet representation?'' We explicitly group the set of these properties into two categories: textual and social. Textual category includes properties such as tweet length, the order of words in it, words, slang words, hashtags, named entities, and capitalization in the tweet. On the other hand, properties such as mention count, first mention position, is reply, reply time and repeating word from a conversation fall under the social category. We investigate the degree to which the tweet representations encode these properties. We assume that \textit{if we cannot train a classifier to predict a property based on its tweet representation, then this property is not encoded in this representation}. For example, the model which preserves the tweet length should perform well in predicting the length given the representation generated from the model. Though these elementary property prediction tasks are not directly related to any downstream application, knowing that the model is good at modeling a particular property (e.g., social properties) indicates that it could excel in correlated applications (e.g., user profiling). In this work we perform an extensive evaluation of nine unsupervised and four supervised tweet representation models, using 13 different properties.

Our main contributions are summarized below.
\begin{itemize}
\item Our work is the first towards fine-grained interpretation of tweet embeddings. To this end, we propose a set of 13 tweet-specific elementary property prediction tasks which help in unearthing both the textual as well as social aspects of different tweet representations.
\item We perform extensive comparison of 13 different various tweet representations with respect to such properties across two dimensions: tweet length and sensitivity to representation size.
\item We draw interesting insights by correlating the model performance obtained for elementary property prediction tasks with multiple downstream applications.
\item Extensive experiments show that bi-directional LSTMs and Skip-Thought vectors (STV) best encode the textual and social properties of tweets respectively. Paragraph2Vec performs the worst while FastText performs best when embedding size needs to be very small.
\end{itemize}
The paper is organized as follows. Section~\ref{sec:related} presents the related work. Sections~\ref{sec:aux_tasks} and~\ref{sec:rep_model} discuss the set of proposed elementary property prediction tasks and the models considered for this study. Sections~\ref{sec:experiments} and~\ref{sec:resultsAnalysis} present the experiment setup and result analysis respectively. We conclude the work with a brief summary in Section~\ref{sec:conclusion}.

\section{Related Work} 
\label{sec:related}

\begin{table*}
\centering
\scriptsize
\caption{Survey of Supervised Representation Learning Models}
\begin{tabular}{|p{2cm}|p{3.2cm}|p{6.5cm}|p{5cm}|}
\hline
\textbf{Model} & \textbf{Architecture} & \textbf{Core Idea} & \textbf{Applications Considered in the Paper} \\ \hline
CNN~\cite{kim14_emnlp} & Simple CNN & Classify using a CNN on top of pre-trained word vectors & Sentiment analysis, Question classification \\
\hline
Tree-LSTM~\cite{tai15_acl} & Recursive Network & Generalization of LSTMs to model recursive nature of sentences & Semantic relatedness, Sentiment classification \\ \hline
FastText~\cite{joulin16_arxiv} & Simple Feed Forward Network & Classify using the average of word vectors & Sentiment analysis, Tag prediction \\ \hline
\end{tabular}
\label{tab:relatedSup}
\end{table*}


\subsection{Summary of Existing Models}
Tables~\ref{tab:relatedUnSup} and~\ref{tab:relatedSup} summarize the core idea and the architecture of the existing unsupervised and supervised models respectively.  Based on the network architecture, neural network models can be classified into one or more of the following categories: Feed Forward, Word2Vec, Encoder-Decoder, Siamese, Convolutional Neural Network (CNN), Recurrent Neural Network (RNN), and Recursive Neural Network (ReNN). 
ReNNs work with parse trees, and hence are ill-suited for representing tweets, as parse tree construction is not only computation intensive but also expects the input sentences to be grammatically well-formed unlike most tweets. Hence we do not consider ReNNs for our study. 

\subsection{Understanding Sentence Representations}
In a recent work, Hill et al.~\cite{hill16_naacl} perform a comparison of different sentence representation models by evaluating them for different high-level semantic tasks such as paraphrase identification, sentiment classification, etc. Our work is different from their work in two ways: (1) They analyze sentences while we work with tweets. Naturally, they ignore social aspects. (2) They survey representations and their effectiveness for various applications; while we perform analysis of representations, and try to estimate their effectiveness for various applications.
The most relevant work to ours is that of~\cite{adi16_arxiv}, which investigates three sentence properties in comparing two models: average of words vectors and LSTM auto-encoders. Our work differs from their work in two ways: (1) While they focus on sentences, we focus on tweets which opens up the challenge of understanding how well these representations capture multiple tweet-specific salient textual properties like slang words, hashtag and unreliable capitalization, and social properties like mentions and conversations. (2) While they work with only 3 properties for 2 models, we provide a more comprehensive analysis by considering 13 properties for 13 different models.

\section{Elementary Property Prediction Tasks} 
\label{sec:aux_tasks}

\begin{table}[]
\centering
\begin{tabular}{|p{0.8in}|p{2in}|p{0.5in}|}
\hline
\textbf{Task} & \textbf{Dataset name} & \textbf{Dataset size} \\ \hline
Length & Sentiment140~\cite{go09_stanford} & 1,98,440 \\ \hline
Content & Sentiment140~\cite{go09_stanford} & 1,98,083 \\ \hline
Word Order & Sentiment140~\cite{go09_stanford} & 1,94,720 \\ \hline
Slang Words & \url{https://noisy-text.github.io/norm-shared-task.html}  & 3,120 \\ \hline
Hashtag & User Profiling~\cite{li14_acl} & 2,00,000 \\ \hline
Named Entities & Twitter NER~\cite{ritter11_emnlp} & 2,394 \\ \hline
Cap. Count & User Profiling~\cite{li14_acl} & 2,00,000 \\ \hline
Informative Cap. & Twitter NER~\cite{ritter11_emnlp} & 400 \\ \hline
Mention Count & User Profiling~\cite{li14_acl} & 2,00,000 \\ \hline
Mention Position & User Profiling~\cite{li14_acl} & 2,00,000 \\ \hline
Is Reply & Conversation~\cite{ritter10_acl} & 75,008 \\ \hline
Reply Time & Conversation~\cite{ritter10_acl} & 31,669 \\ \hline
Word Repetition in Conversation &  Conversation~\cite{ritter10_acl} & 37,504 \\ \hline
\end{tabular}
\caption{Dataset Statistics}
\label{tab:data}
\end{table}

In this section we list down the set of proposed elementary property prediction tasks to test the characteristics of a tweet embedding. These properties correspond to the most popular features used in multiple papers using feature engineering for various microblog applications. Since tweets are pieces of text in a network context, we naturally categorized the properties into two types: \textit{textual} and \textit{social}. Note that we use a neural network to build the elementary property prediction task classifier which has the following two layers in order: the representation layer, and the softmax layer on top whose size varies according to the specific task. When there are more than one input for a task, we concatenate embeddings for each input. Table~\ref{tab:data} presents the dataset statistics for each task. 


\subsection{Textual Tasks}
Unlike sentences, tweets have slang words, entities, hashtags, unreliable capitalization, etc. We evaluate tweet representations against the following tasks to check their robustness against this noise. 

\noindent\textbf{(a) \underline{Length Task}}: Tweet length is a useful feature for detecting spam tweets, news-worthy tweets, etc. This task measures the extent to which the tweet representation encodes the length of the tweet. Given a tweet embedding, the task is to predict the number of words in the tweet. We use binned length to do multi-class classification. After varying bin size in a reasonable range (3--6), we did not observe much change in the results, hence we show results for bin size set as 4.

\noindent\textbf{(b) \underline{Content Task}}: Words in a tweet is a useful feature for sentiment analysis, paraphrase detection, response prediction, etc. This task measures the extent to which the tweet representation encodes the identities of words present in it. Given a tweet embedding and a word embedding, the task is to predict whether the word is in the tweet or not. This is posed as binary classification task where we inject randomly selected words not appearing in the tweet to generate negative samples.

\noindent\textbf{(c) \underline{Word Order Task}}: Word order is a useful feature in textual tasks like parsing. This task measures the extent to which the tweet representation preserves the word order. Given a tweet embedding, the embeddings of two words, $\it{w_1}$ and $\it{w_2}$ that appear in the tweet, the task is to predict whether the word $\it{w_1}$ appears before the word $\it{w_2}$ in the tweet or not. This is solved as a binary classification task, where the order of words are flipped to generate negative samples.

\noindent\textbf{(d) \underline{Slang Words Task}}: Slang word is a useful feature in tasks such as sentiment analysis, paraphrase detection, etc. This task measures the extent to which the tweet representation is robust to the non-standard spellings (e.g., `toook' for `took'), informal abbreviations (e.g., `tmrw' for `tomorrow'), etc., which are ubiquitous on Twitter. Given a tweet embedding, and the embeddings of two words ($\it{w_1}, \it{w_2}$), the task is to predict whether the word $\it{w_2}$ is the canonical form of the word $\it{w_1}$ (which is present in the tweet) or not. This is also posed as a binary classification task, where the word $\it{w_2}$ is randomly sampled to generate negative samples. 

\noindent\textbf{(e) \underline{Hashtag Task}}: Hashtag is a useful feature in tasks such as sentiment analysis, hashtag prediction, response prediction, etc. This task measures the extent to which the tweet representation encodes the identities of hashtags present in the tweet. Given a tweet embedding and an embedding of the word that appears in the tweet, the task is to predict whether the word is a hashtag or not. This is solved as a binary classification task, where the negative samples are generated by randomly sampling words from the tweet which are not hashtags. 

\noindent\textbf{(f) \underline{Named Entity (NE) Task}}: Named entities are a useful feature in detecting paraphrases, etc. This task measures the extent to which the tweet representation encodes the identities of the named entities present in the tweet. Given a tweet embedding and an embedding of the n-gram that appears in the tweet, the task is to predict whether the n-gram is a NE or not. This is solved as a binary classification task, where the negative samples are generated by randomly sampling n-grams from the tweet which are not NEs. 

\noindent\textbf{(g) \underline{Capitalization Count Task}}: Capitalization count is a useful feature in detecting named entities, paraphrases and so on. This task measures the extent to which the tweet representation encodes the number of capitalized words present in the tweet. Given a tweet embedding, the task is to predict the number of words starting with a capital letter in the tweet. 

\noindent\textbf{(h) \underline{Informative Capitalization Task}}: Capitalization is a key orthographic feature for recognizing NE. Unlike in curated text, non-entity words in some tweets are capitalized just for emphasis and could confuse a na\"{\i}ve named entity recognizer. In this task, we measure the extent to which the tweet representation encodes the capitalized word which are informative in identifying the named entity mention. Given a tweet embedding and an embedding of the capitalized word that appears in the tweet, the task is to predict whether the word in the tweet is informative for identifying NE mention or not. This is also framed as a binary classification task. 


\subsection{Social Tasks}
Besides the textual properties, a good representation should be able to explain the following social properties of tweets.

\noindent\textbf{(i) \underline{Mention Count Task}}: Mention count is a useful feature in tasks such as sentiment analysis, response prediction, etc. This task measures the extent to which the tweet representation encodes the number of mentions present in it. Given a tweet embedding, the task is to predict the number of user mentions (words starting with the letter `@') in the tweet. We use the raw frequency and pose it as a classification problem.

\noindent\textbf{(j) \underline{Mention Position Task}}: Mention position is a useful feature in tasks such as sentiment analysis, response prediction, etc. This task measures the extent to which the representation encodes the position of the first user mention in the tweet. 

\noindent\textbf{(k) \underline{Is Reply Task}}: This task measures the extent to which the tweet representation encodes the salient properties of a reply tweet. Given a tweet embedding, the task is to predict whether the tweet is a reply tweet or not. To generate the negative instances for this binary task, we randomly choose a tweet that is a conversation starter.

\noindent\textbf{(l) \underline{Reply Time Task}}: Reply time is a useful feature in modeling the conversation, predicting responses, etc. This task measures the extent to which the tweet representation encodes the temporal aspects of a reply tweet. Given a tweet embedding, the task is to predict the number of minutes taken to get a reply for the tweet. For simplicity, we consider only the tweets which get a reply within an hour. 

\noindent\textbf{(m) \underline{Word Repetition in Conversation Task}}: Word repetition in a conversation is a useful feature in modeling the conversation, predicting responses, etc. This task measures the extent to which the tweet representation encodes the frequent words in a conversation. Given a tweet embedding and an embedding for a word, the task is to predict whether the word will be used the most in the ensuing conversation thread from the tweet that is a conversation starter. We randomly choose the word that is never used later in the conversation in order to generate negative samples.


\section{Representation Learning Models} 
\label{sec:rep_model}

In this section we list down popular models for learning tweet representations. 
\subsection{Unsupervised Models}
We experiment with the following unsupervised representation learning models. These models require an additional classifier to do the final classification.
\begin{itemize}
\item \textbf{Bag Of Words} (BOW)~\cite{wang16_acl} - This simple representation captures the TF-IDF value of an n-gram. We pick top 50K n-grams, with the value of `n' going upto 5.
\item \textbf{Latent Dirichlet Allocation} (LDA)~\cite{wang16_acl} - We use the topic distribution resulting by running LDA with number of topics as 200, as the tweet representation. We varied number of topics as 100, 200, 500 but found best results at 200.
\item \textbf{Bag Of Means} (BOM) - We take the average of the word embeddings obtained by running the GloVe~\cite{pennington14_emnlp} model on 2B tweets with embedding size as 200~\footnote{http://nlp.stanford.edu/projects/glove/}. We varied embedding size as 25, 50, 100, 200 (since these are the available sizes) but found best results at 200.
\item \textbf{Deep Structured Semantic Models} (DSSM)~\cite{huang13_cikm} - This is a deep encoder trained to represent query and document in common space, for the document ranking task. We use the publicly available pre-trained encoder to encode the tweets~\footnote{https://www.microsoft.com/en-us/research/project/dssm/}.
\item \textbf{Convolutional DSSM} (CDSSM)~\cite{shen14_cikm} - This is the convolutional variant of DSSM.
\item \textbf{Paragraph2Vec} (PV)~\cite{le14_icml} - This model based on Word2Vec~\cite{mikolov13_nips} learns embedding for a document which is good in predicting the words within it. We use the BOW variant with the recommended embedding size and window size of 200 and 10 respectively.
\item \textbf{Skip-Thought Vectors} (STV)~\cite{kiros15_nips} - This is a Gated Recurrent Unit (GRU) encoder trained to predict adjacent sentences in a books corpus. We use the recommended combine-skip (4800-dimensional) vectors from the publicly available encoder~\footnote{https://github.com/ryankiros/skip-thoughts}.
\item \textbf{Tweet2Vec} (T2V)~\cite{dhingra16_acl} - This is a character composition model working directly on the character sequences to predict the user-annotated hashtags in a tweet. We use publicly available encoder, which was trained on 2M tweets~\footnote{https://github.com/bdhingra/tweet2vec}.
\item \textbf{Siamese CBOW} (SCBOW)~\cite{kenter16_acl} - This model uses averaging of word vectors to represent a sentence, and the objective and data used here is same as that for STV. Note that this is different from BOW because the word vectors here are optimized for sentence representation.
\end{itemize}
\subsection{Supervised Models}
Below we list the set of supervised representation learning models which we use for end-to-end classification.
\begin{itemize}
\item \textbf{Convolutional Neural Network} (CNN) - This is a simple CNN proposed in~\cite{kim14_emnlp}.
\item \textbf{Long Short Term Memory Network} (LSTM)~\cite{graves13_icassp} - This is a vanilla LSTM based recurrent model, applied from start to the end of a tweet, and the last hidden vector is used as tweet representation. We use the optimal hyper-parameter settings as proposed in~\cite{tai15_acl}.
\item \textbf{Bi-directional LSTM} (BLSTM)~\cite{graves13_icassp} - This extends LSTM by using two LSTM networks, processing a tweet left-to-right and right-to-left respectively. A tweet is represented by concatenating the last hidden vector of both LSTMs. We use the optimal hyper-parameter settings proposed in~\cite{tai15_acl}.
\item \textbf{FastText} (FT)~\cite{joulin16_arxiv} - This is a simple architecture which averages the n-gram vectors to represent a tweet, followed by the softmax in the final layer. 
\end{itemize}

\section{Experiments} 
\label{sec:experiments}
\begin{table*}
\centering
\scriptsize
\caption{Detailed Analysis of Unsupervised and Supervised Models}
\begin{tabular}{|p{0.2cm}|p{1cm}|p{3.3cm}|p{6.8cm}|p{6.8cm}|} 
\hline
& \textbf{Model} & \textbf{Task Accuracy (1)} & \textbf{Tweet Length (2)} & \textbf{Representation Size (3)} \\ \hline
\multirow{9}{*}{\begin{sideways}\parbox{5cm}{\textbf{Unsupervised}}\end{sideways}} & BOW & 
(a): Reply Time \newline
(b): Slang Words, Reply Time 
&
(a): Is Reply, Reply Time, Word Repeat \newline
(b): Length, Word Order, Capt. Count, Mention Count, Mention Pos.
&
(a): Word Repeat \newline
(b): Content, Word Order, Slang Word, NE, Capt. Count, Mention Count, Mention Pos., Is Reply, Reply Time \newline
(c): Length, Hashtag
\\  
\cline{2-5}
\cline{2-5}
& LDA &
(a): Hashtag \newline
(b): Hashtag
&
(a): Is Reply, Reply Time, Word Repeat
&
(a): Length, Hashtag, Capt. Count, Reply Time \newline
(b): Content, Word Order, Slang Word, NE, Info. Cap., Mention Count, Mention Pos., Word Repeat \newline
(c): Is Reply
\\ \cline{2-5}
& BOM &
(a): Word Order, NE  \newline
(b): Word Order, NE
&
(a): Is Reply, Reply Time, Word Repeat \newline
(b): Length, Word Order, Capt. Count, Mention Count, Mention Pos.
&
(a): Content, Word Order, Hashtag, NE, Word Repeat \newline
(b): Length, Capt. Count, Mention Count, Mention Pos, Is Reply
\\ \cline{2-5}
& DSSM &
Best in none &
(a): Is Reply, Reply Time, Word Repeat \newline
(b): Length, Word Order, Capt. Count, Mention Count, Mention Pos.
&
Used pre-trained embeddings and so did not study this effect.
 \\ \cline{2-5}
& CDSSM &
(a): Hashtag \newline
(b): Hashtag &
(a): Is Reply, Reply Time, Word Repeat \newline
(b): Length, Word Order, Capt. Count, Mention Count, Mention Pos.
&
Used pre-trained embeddings and so did not study this effect.
\\ \cline{2-5}
& PV &
Best in none &
(a): Is Reply, Reply Time, Word Repeat \newline
(b): Length, Word Order, Capt. Count, Mention Count, Mention Pos.
&
Used pre-trained embeddings and so did not study this effect.
 \\ \cline{2-5}
& STV &
(a): Content, Mention Count, Is Reply, Word Repeat \newline
(b): Content, Capt. Count, Mention Count, Mention Pos., Is Reply, Word Repeat &
(a): Is Reply, Reply Time, Word Repeat \newline
(b): Length, Word Order, Capt. Count, Mention Count, Mention Pos.
&
Used pre-trained embeddings and so did not study this effect.
\\ \cline{2-5}
& T2V &
(b): Length, Info. Capt &
(a): Is Reply, Reply Time, Word Repeat  \newline
(b): Length, Word Order, Capt. Count, Mention Pos.
&
Used pre-trained embeddings and so did not study this effect.
 \\ \cline{2-5}
& SCBOW &
(a): Hashtag \newline
(b): Hashtag &
(a): Is Reply, Reply Time, Word Repeat \newline
(b): Length, Word Order, Capt. Count, Mention Count, Mention Pos.
&
Used pre-trained embeddings and so did not study this effect.
\\ \hline
\multirow{4}{*}{\begin{sideways}\parbox{2.5cm}{\textbf{Supervised}}\end{sideways}}  & CNN & 
(a): Info. Capt \newline
(c): Content, Word Order, Hashtag, NE, Info. Capt, Mention Count, Reply Time &
(a): Is Reply, Reply Time, Word Repeat \newline
(b): Length, Word Order, Capt. Count, Mention Count, Mention Pos.
&
(b): Length, Content, Word Order, Slang Words, Hashtag, NE, Capt. Count, Mention Count, Reply Time, Word Repeat, Mention Pos.
  \\  
\cline{2-5}
\cline{2-5}
& LSTM &
(a): Length \newline
(c): Length &
(a): Is Reply, Reply Time, Word Repeat \newline
(b): Word Order, Capt. Count, Mention Count, Mention Pos.
&
(b): Content, Word Order, Slang Words, Hashtag, NE, Capt. Count, Mention Count, Is Reply, Reply Time, Word Repeat, Mention Pos. 
 \\ \cline{2-5}
& BLSTM &
(a): Slang Words, Capt. Count, Mention Pos. \newline
(c): Slang Words, Capt. Count, Mention Pos., Is Reply &
(a): Is Reply, Reply Time, Word Repeat \newline
(b): Word Order, Capt. Count, Mention Pos.
&
(b): Content, Word Order, Slang Words, Hashtag, NE, Capt. Count, Mention Count, Reply Time, Word Repeat, Mention Pos.
\\ \cline{2-5}
& FastText &
(c) Word Repeat &
(a): Is Reply, Reply Time, Word Repeat \newline
(b): Length, Word Order, Capt. Count, Mention Pos.
&
(a): Length, Slang Words, Hashtag, NE, Capt. Count, Info. Capt, Mention Count, Is Reply, Word Repeat, Mention Pos. \newline
(b): Content
 \\ \hline
\end{tabular}
\label{tab:analysis}
\end{table*}

\begin{table*}
\centering
\scriptsize
\caption{Elementary Property Prediction Task F1-Score (\%) - Performance Comparison}
\begin{tabular}{|l|p{1cm}|p{0.7cm}|p{0.8cm}|p{0.7cm}|p{0.7cm}|p{0.7cm}|p{0.7cm}|p{1.5cm}|p{1.5cm}|p{1cm}|p{1cm}|p{0.7cm}|p{0.7cm}|p{2cm}|}
\hline
 & \textbf{Model / Task} & \textbf{Length} & \textbf{Content} & \textbf{Word Order} & \textbf{Slang Words} & \textbf{Hash tag} & \textbf{Named Entity} & \textbf{Capitalization Count} & \textbf{Informative Capitalization} & \textbf{Mention Count} & \textbf{Mention Position} & \textbf{Is\ \ \  Reply} & \textbf{Reply Time} & \textbf{Word Repetition in Conversation} \\ \hline
\multirow{9}{*}{\begin{sideways}\textbf{Unsupervised}\end{sideways}} & 
BOW & 37.83 & 97.37 & 60.36 & 78.13 & 99.28 & 89.66 & 37.59 & 72.39 & 87.02 & 75.2 & 78.14 & \textbf{35.98} & 86.92   \\  
\cline{2-15}
\cline{2-15}
& LDA & 25.11 & 97.72 & 60.62 & 76.82 & \textbf{99.35} & 97.24 & 55.25 & 68.66 & 69.31 & 36.85 & 60.12 & 28.03 & 91.71 \\ \cline{2-15}
& BOM & 47.64 & 98.67 & \textbf{61.25} & 75.26 & 99.33 & \textbf{98.06} & 59.1 & 73.13 & 74.26 & 45.64 & 66.25 & 28.43 & 92.26 \\ \cline{2-15}
& DSSM & 57.76 & 98.57 & 59.01 & 76.89 & 99.33 & 97.16 & 69.57 & 71.64 & 78.86 & 52.46 & 76.47 & 29.08 & 91.93 \\ \cline{2-15}
& CDSSM & 47.75 & 98.09 & 57.66 & 69.8 & \textbf{99.35} & 97.41 & 62.42 & 68.66 & 75.48 & 44.38 & 73.92 & 28.49 & 92.47 \\ \cline{2-15}
& PV & 13.58 & 94.9 & 60.92 & 76.09 & 85.61 & 98.02 & 33.14 & 70.89 & 45.98 & 21.3 & 54.68 & 27.58 & 90.71 \\ \cline{2-15}
& STV & 71.85 & \textbf{98.85} & 57.7 & 76.66 & 99.32 & 97.92 & 72.28 & 70.89 & \textbf{98.94} & 68.46 & \textbf{96.41} & 29.25 & \textbf{92.82} \\ \cline{2-15}
& T2V & 73.58 & 98.36 & 60.62 & 62.34 & 99.32 & 92.93 & 71.81 & 82.84 & 86.5 & 66.29 & 95.73 & 31.59 & 89.76 \\ \cline{2-15}
& SCBOW & 32.13 & 97.94 & 58.39 & 74.24 & \textbf{99.35} & 97.79 & 43.32 & 69.4 & 70.62 & 33.59 & 60.38 & 28.39 & 92.49 \\ \hline
\multirow{4}{*}{\begin{sideways}\parbox{1.1cm}{\textbf{Supervised}}\end{sideways}} & 
CNN & 59.48 & 97.71 & 61.13 & 77.42 & 99.31 & 91.38 & 86.32 & \textbf{88.81} & 91.04 & 77.15 & 92.66 & 31.73 & 87.28 \\ \cline{2-15}
& LSTM & \textbf{99.79} & 97.39 & 60.74 & 76.24 & 99.28 & 90.36 & 89.23 & 61.19 & 89.49 & 80.26 & 92.39 & 28.46 & 85.58 \\ \cline{2-15}
& BLSTM & 98.72 & 97.47 & 60.85 & \textbf{80.52} & 99.28 & 90.89 & \textbf{89.73} & 75.37 & 89.83 & \textbf{91.2} & 92.76 & 27.99 & 86.43 \\ \cline{2-15}
& FastText & 24.56 & 92.15 & 60.06 & 67.48 & 89.11 & 78.89 & 76.14 & 57.46 & 83.19 & 61.31 & 74.08 & 28.35 & 70.58 \\ \hline
\end{tabular}
\label{tab:ta}
\end{table*}

We perform an extensive evaluation of all the models in an attempt to find the significance of different representation models. Essentially we study every model (with optimal settings reported in the corresponding paper) with respect to the following perspectives.

\renewcommand{\theenumi}{(\alph{enumi}}

\noindent\textbf{(1) Property prediction task accuracy} - When the accuracy of the model for a property is high, it is more likely to encode the property. This test identifies the model with the best F1-score for each elementary property prediction task. 
	\begin{enumerate}
	\item \textit{Best model}: Tasks for which this model has outperformed all the other models.
	\item \textit{Best unsupervised model}: Tasks for which this model has outperformed all the other unsupervised models.
	\item \textit{Best supervised model}: Tasks for which this model has outperformed all the other supervised models.
	\end{enumerate}
\noindent\textbf{(2) Property prediction task accuracy versus Tweet length} - Some representation learning models are biased towards modeling shorter or longer tweets. This test helps to compare the performance of the model for shorter vs. longer tweets.
	\begin{enumerate}
	\item \textit{Positively correlated tasks}: Tasks for which the performance of the model increases as tweet length increases.
	\item \textit{Negatively correlated tasks}: Tasks for which the performance of the model decreases as tweet length increases.
	\end{enumerate}
\noindent\textbf{(3) Property prediction task accuracy versus Representation size} - Embedding size is an important hyper-parameter to tune the performance of the representation learning model. This test captures the sensitivity of each model with respect to the embedding size. 
	\begin{enumerate}
	\item \textit{Invariant tasks}: Tasks for which the model performance is invariant with increase in embedding size.
	\item \textit{Positively correlated tasks}: Tasks for which the model performance increases with increase in embedding size.
	\item \textit{Negatively correlated tasks}: Tasks for which the model performance decreases with increase in embedding size.
	\end{enumerate}

\section{Results and Analysis} 
\label{sec:resultsAnalysis}

Detailed analysis of various supervised and unsupervised models discussed in Section~\ref{sec:rep_model}, across various dimensions (1, 2, and 3) discussed in Section~\ref{sec:experiments}, is presented in Table~\ref{tab:analysis}. Thus, e.g., the ``task accuracy (1)'' column entry for STV tells us that STV is ``best model (a)'' for Content, Mention Count, Is Reply, Word Repeat tasks, and ``best unsupervised model (b)'' for Content, Capt. Count, Mention Count, Mention Pos., Is Reply, Word Repeat tasks. We discuss these in detail in this section. We have made the code publicly available at \url{http://tinyurl.com/mysteriousTweetReps}. In this section, we analyze the results in detail.

\subsection{Property Prediction Task Accuracy}
\label{a:task_acc}
We summarize the results of all the property prediction tasks in Table~\ref{tab:ta}. 

For the textual tasks, we observe the following. (1) Length prediction turns out to be a difficult task for most of the models. Models which rely on the recurrent architectures such as LSTM, STV, T2V have sufficient capacity to perform well in modeling the tweet length. PV performs the worst due to the simple Word2Vec structure on this complex task. FastText expectedly loses the length information and performs as the worst supervised model. (2) BLSTM is the best in modeling slang words. BLSTM outperforms the LSTM variant in all the tasks except `Content', which signifies the power of using the information flowing from both the directions of the tweet. (3) STV tops in modeling the content. This is due to the very large embedding size used for every tweet (4800 dimensions). Surprisingly, T2V which is expected to perform well on the `Content' task because of its ability to work at a finer level, i.e., characters performs the worst. In fact T2V does not outperform other models in any task, which could be mainly due to the fact that the hashtags which are used for supervision in learning tweet representations reduce the generalization capability of the tweets beyond hashtag prediction. Prediction tasks such as `Content' and `Hashtag' seem to be less difficult as all the models perform nearly optimal for them. The superior performance of all the models for the `Content' task in particular is unlike the relatively lower performance reported in~\cite{adi16_arxiv}, mainly because of the short length of the tweets. (4) 
BOM performed well on identifying the named entities in the tweet. BOM is raised from the generic word-word statistics, which seems to be just enough for remembering NEs. (5) For the capitalization tasks such as `Capitalization Count' and `Informative Capitalization', we observe the supervised models perform better than the unsupervised models. 

For the social tasks we observe the following. (1) On average, supervised models work better than unsupervised ones on Mention Count, Mention Position and Is Reply tasks. For the other two social tasks, unsupervised models are marginally better. (2) STV is good for most of the social tasks including `Mention Count', `Is Reply' and `Word Repetition'. We believe the main reason for STV's performance is two-fold: (a) the inter-sentential features extracted from STV's encoder by the prediction of the surrounding sentences in the books corpus contains rich social elements that are vital for social tasks (e.g., user profiling), and (b) the recurrent structure in both the encoder and decoder persists useful information in the memory nicely. The second claim is further substantiated by observing the poor performance of SCBOW whose objective is also similar to STV, but with a simpler architecture, i.e., word vector averaging. 

\vspace{-0.138in}
\subsection{Sensitivity to Tweet Length}
\begin{figure}[!htb]
    \centering
    \includegraphics[width=0.8\hsize]{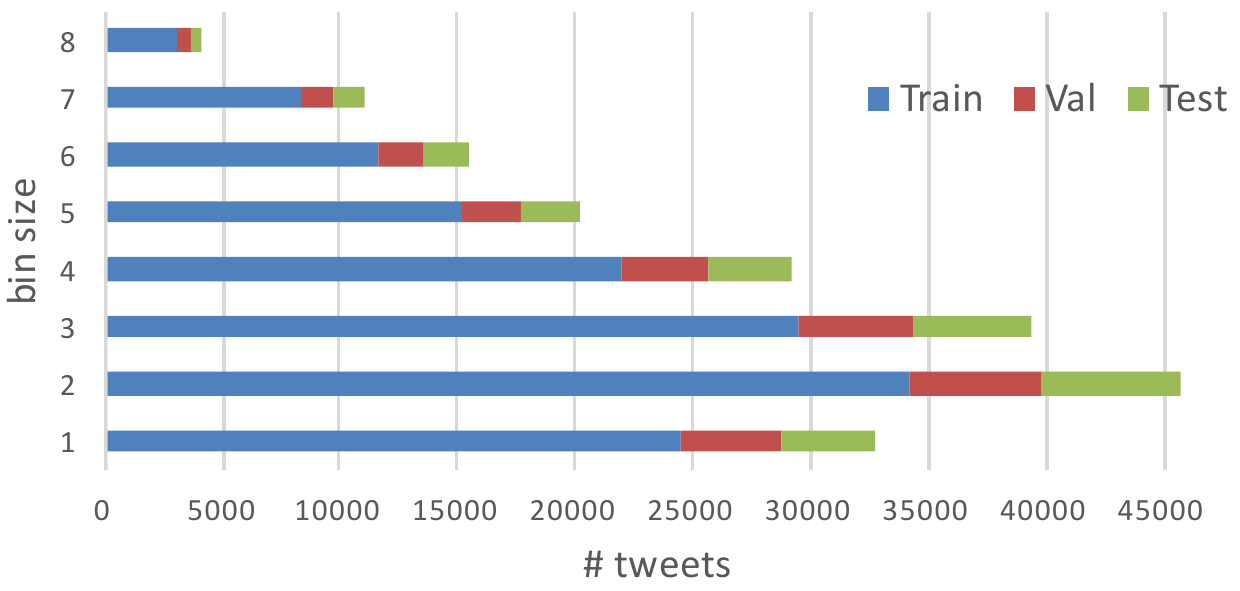}
    \caption{Statistics on Number of Tweets in each Bin of Tweet Length}
    \label{fig:bsize}
\end{figure}

\begin{figure*}[!htb]
    \centering
    \begin{subfigure}[t]{0.3\textwidth}
        \centering
        \includegraphics[height=0.63in,width=2in]{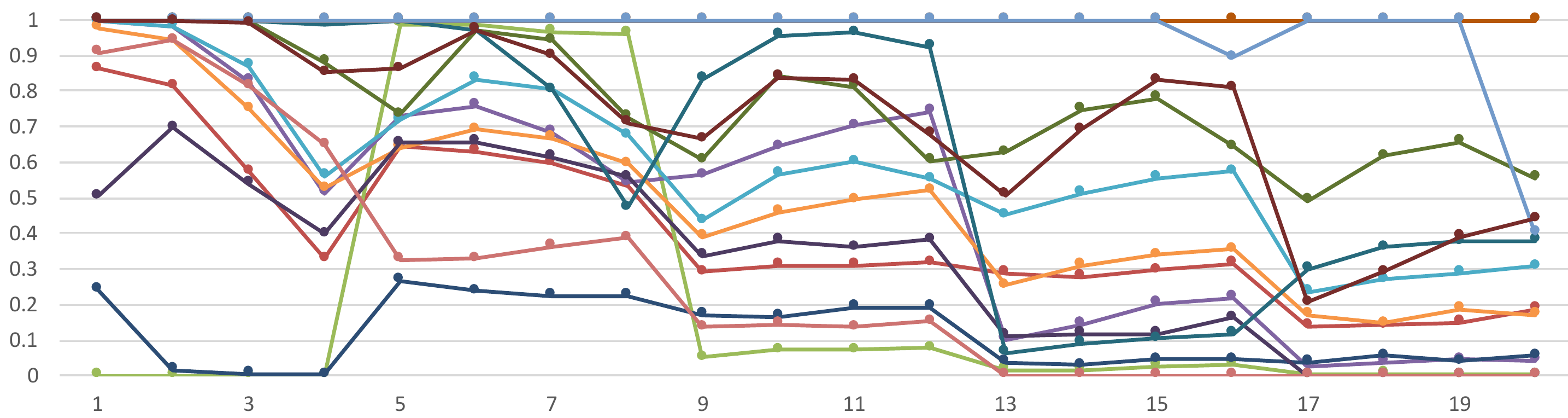}
        \caption{Length}
    \end{subfigure}%
    ~ 
    \begin{subfigure}[t]{0.3\textwidth}
        \centering
        \includegraphics[height=0.63in,width=2in]{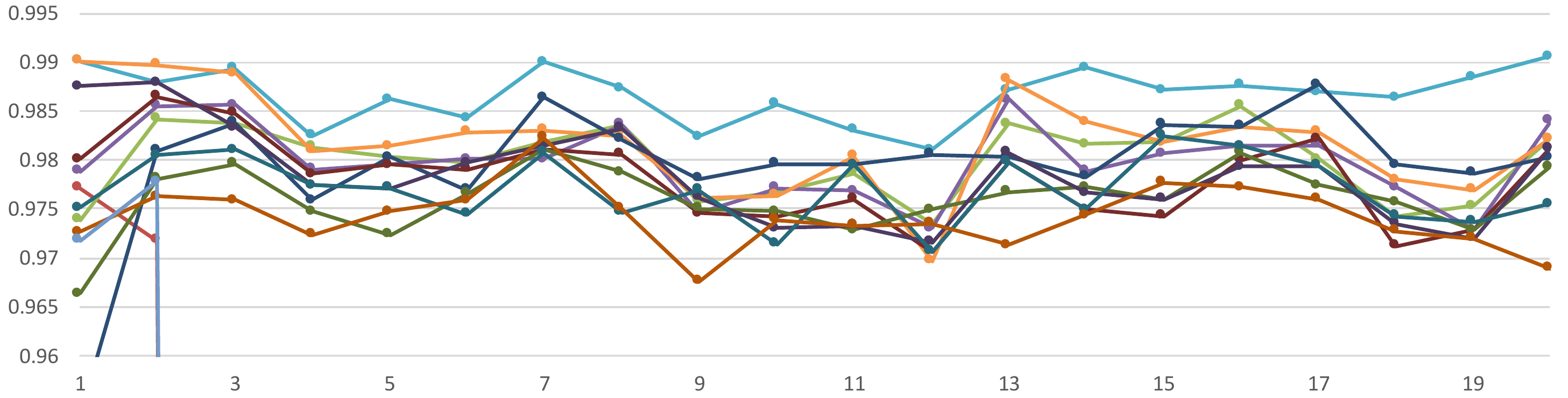}
        \caption{Content}
    \end{subfigure}
    ~
    \begin{subfigure}[t]{0.3\textwidth}
        \centering
        \includegraphics[height=0.63in,width=2in]{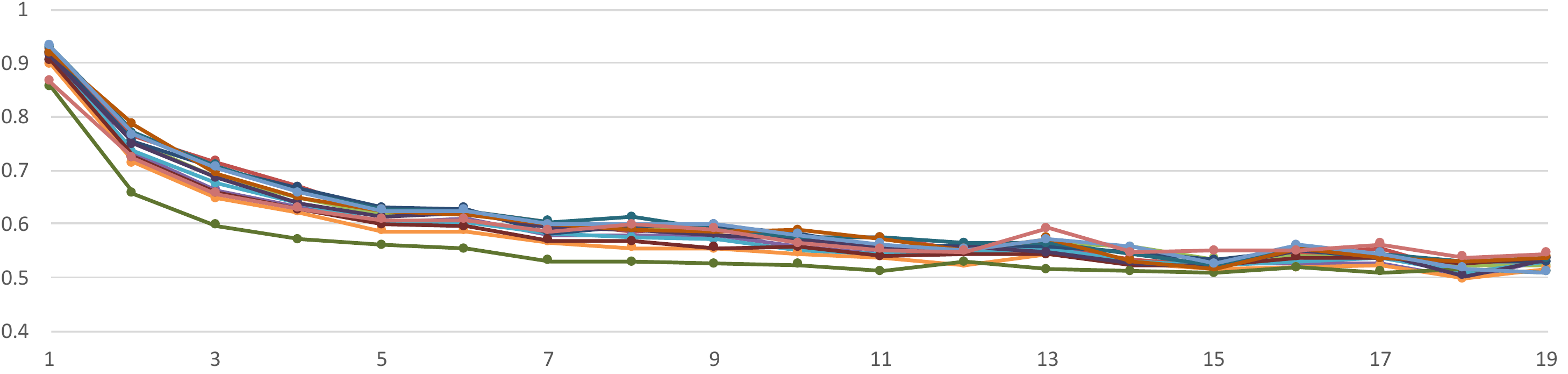}
        \caption{Word Order}
    \end{subfigure}
    
    \begin{subfigure}[t]{0.3\textwidth}
        \centering
        \includegraphics[height=0.63in,width=2in]{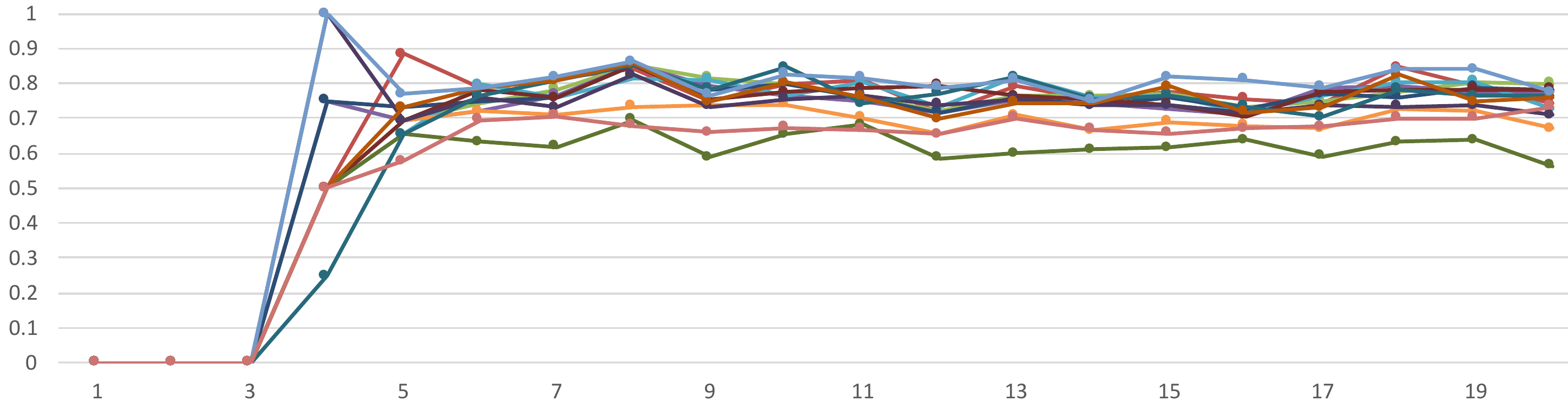}
        \caption{Slang Words}
    \end{subfigure}
    ~
    \begin{subfigure}[t]{0.3\textwidth}
        \centering
        \includegraphics[height=0.63in,width=2in]{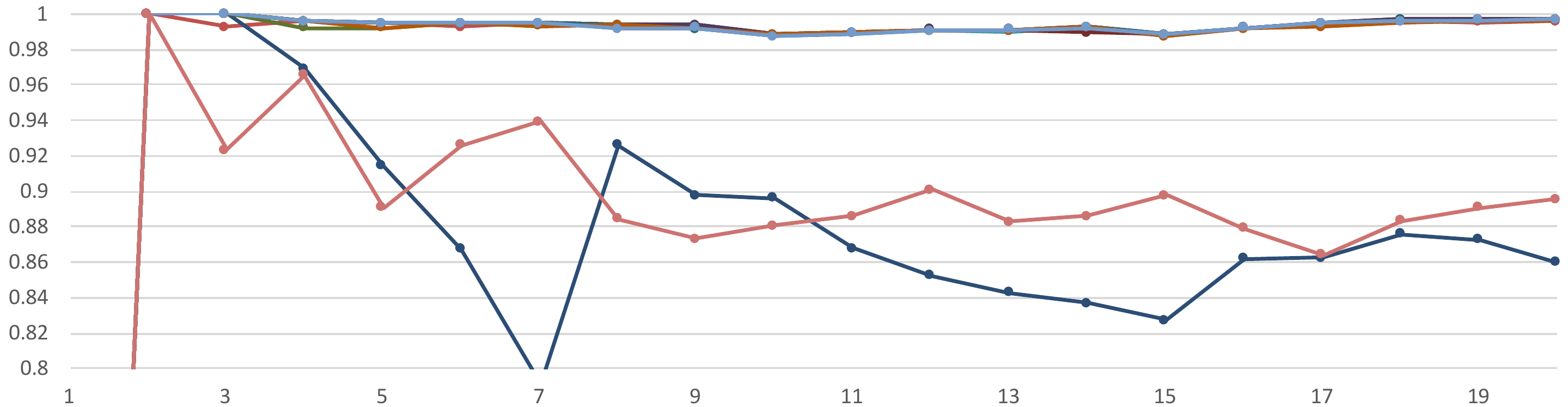}
        \caption{Hashtag}
    \end{subfigure}
    ~
    \begin{subfigure}[t]{0.3\textwidth}
        \centering
        \includegraphics[height=0.63in,width=2in]{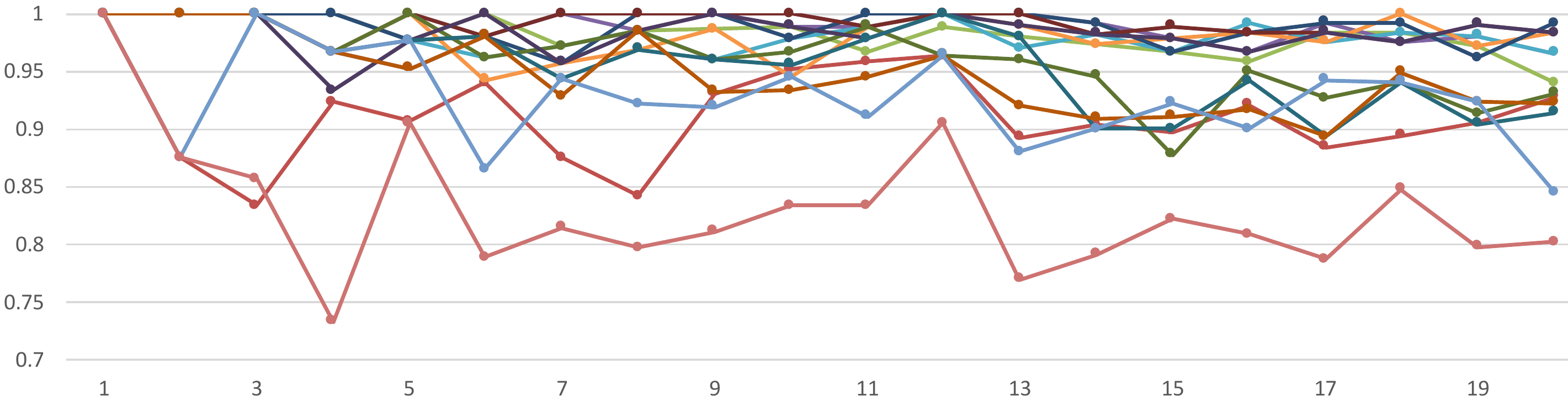}
        \caption{Named Entities}
    \end{subfigure}
    
    \begin{subfigure}[t]{0.3\textwidth}
        \centering
        \includegraphics[height=0.63in,width=2in]{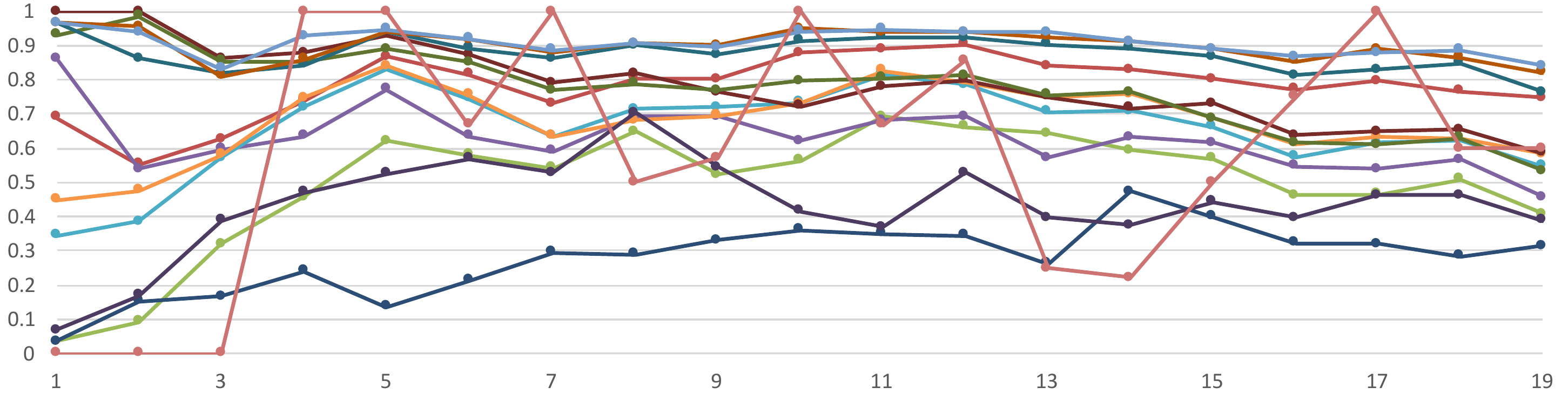}
        \caption{Capitalization Count}
    \end{subfigure}
    ~ 
    \begin{subfigure}[t]{0.3\textwidth}
        \centering
        \includegraphics[height=0.63in,width=2in]{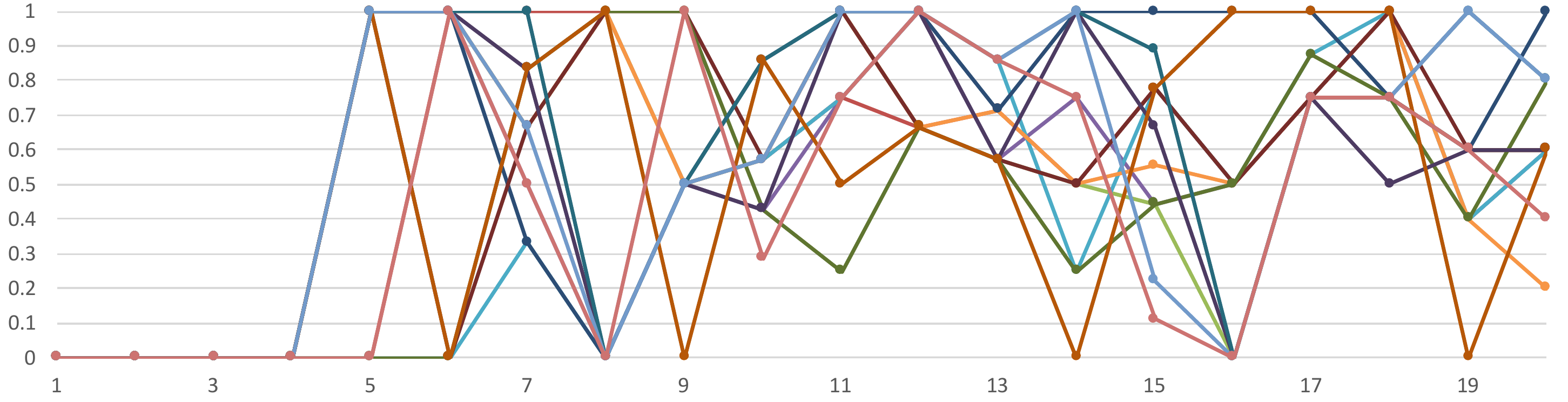}
        \caption{Informative Capitalization}
    \end{subfigure}    
    ~
    \begin{subfigure}[t]{0.3\textwidth}
        \centering
        \includegraphics[height=0.63in,width=2in]{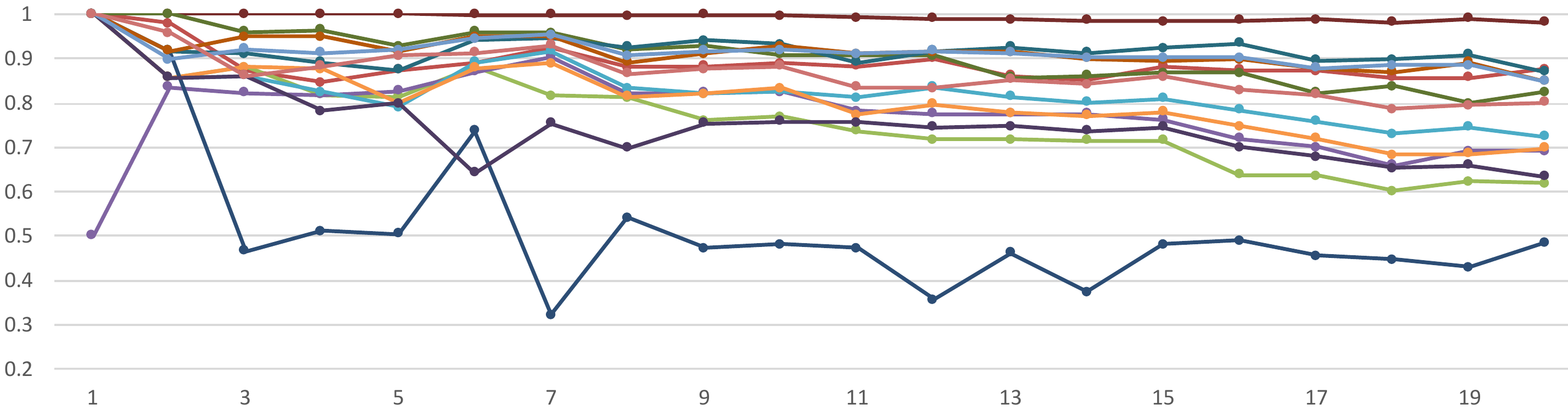}
        \caption{Mention Count}
    \end{subfigure}%
    
    \begin{subfigure}[t]{0.3\textwidth}
        \centering
        \includegraphics[height=0.63in,width=2in]{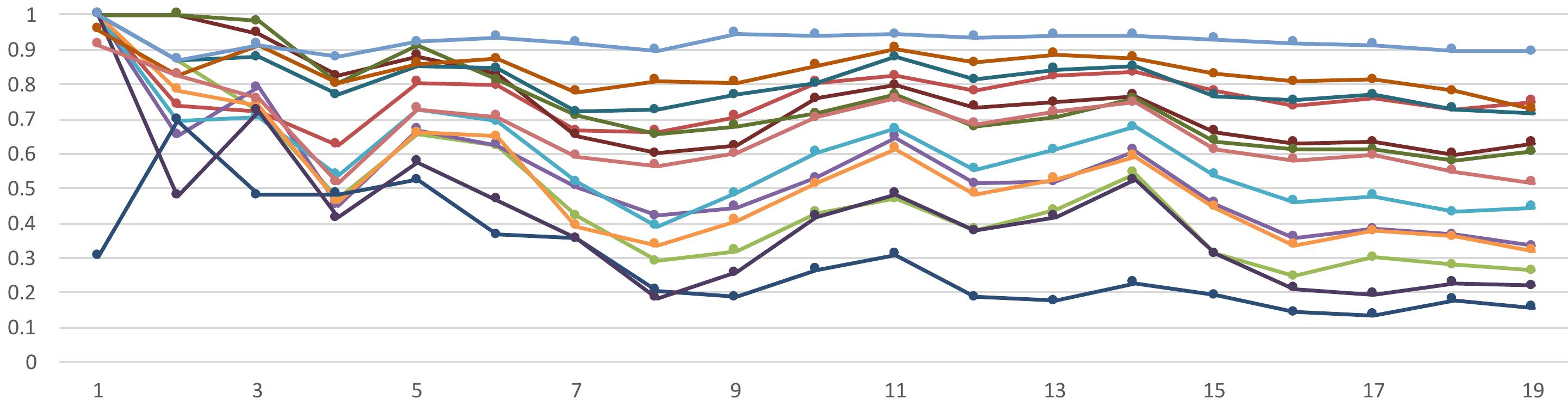}
        \caption{Mention Position}
    \end{subfigure}
    ~
    \begin{subfigure}[t]{0.3\textwidth}
        \centering
        \includegraphics[height=0.63in,width=2in]{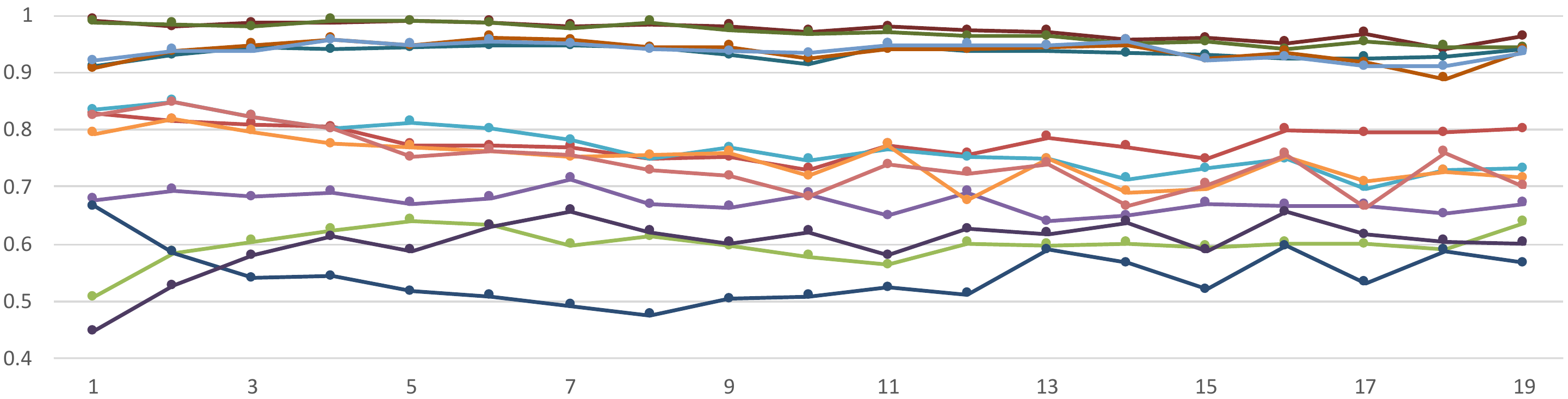}
        \caption{Is Reply}
    \end{subfigure}
    ~
    \begin{subfigure}[t]{0.3\textwidth}
        \centering
        \includegraphics[height=0.63in,width=2in]{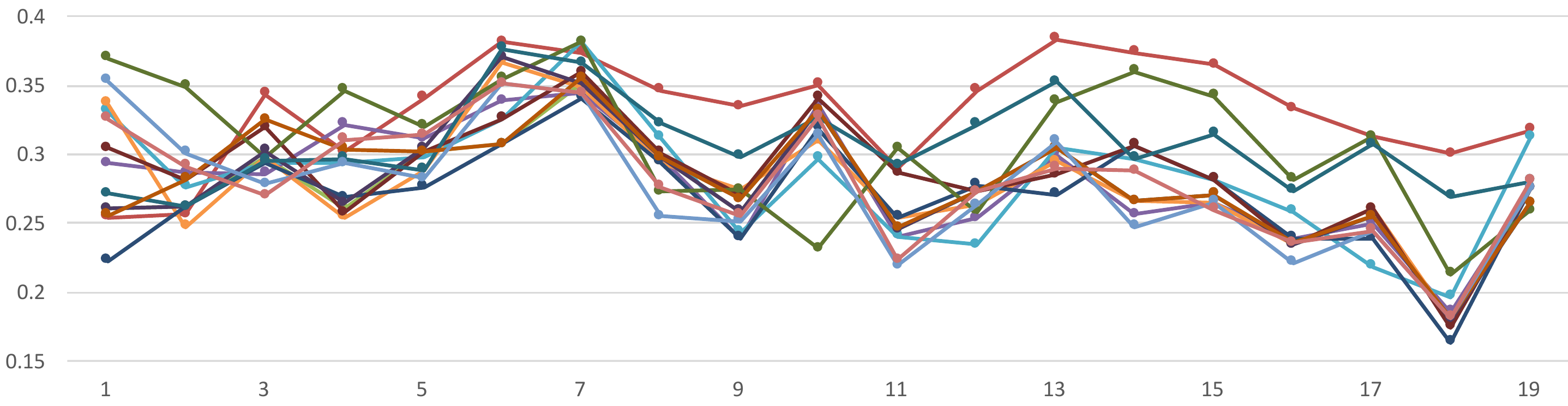}
        \caption{Reply Time}
    \end{subfigure}
    
    \begin{subfigure}[t]{0.3\textwidth}
        \centering
        \includegraphics[height=0.63in,width=2in]{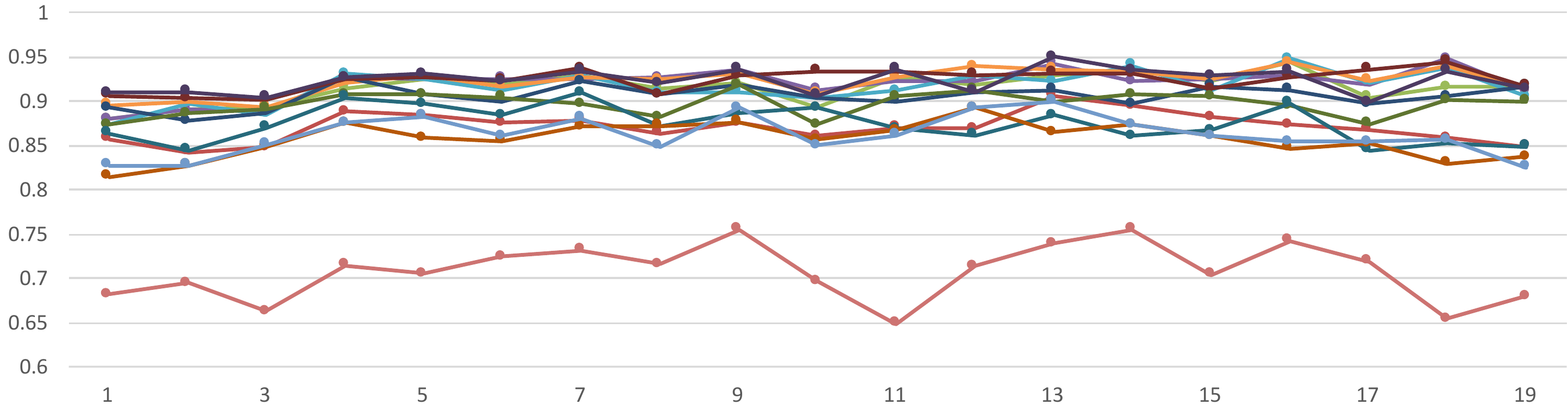}
        \caption{Word Repetition in Conversation}
    \end{subfigure}

    \begin{subfigure}[t]{1\textwidth}
        \centering
        \includegraphics[height=0.15in,width=7in]{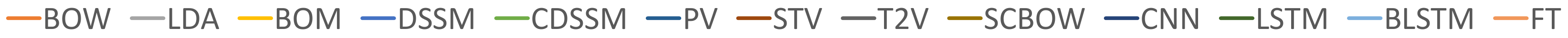}
    \end{subfigure}
    \caption{Performance of the model w.r.t Tweet Length}
    \label{fig:tl}
\end{figure*}

This setup captures the behavior of the model with the increase in the context size, which is defined in terms of number of words. Figure~\ref{fig:bsize} provides the statistics on number of tweets in each bin of tweet length. For tasks such as `Word Order', `Mention Position' and `Capitalization Count', we see the performance of all the models (Figure~\ref{fig:tl}) to be negatively correlated with the tweet length. On the other hand, there is no correlation between the tweet length and the performance of all the models for the tasks such as `Slang Words', `Content', `Hashtag', `Named Entities', `Informative Capitalization' and `Is Reply'. For `Is Reply' task, we see a positive correlation between the tweet length and the performance of all the models (Figure~\ref{fig:tl}). But there is no such correlation for other social tasks such as `Reply Time' and `Word Repetition'. 

\begin{figure*}
    \centering
    \begin{subfigure}[t]{0.2\textwidth}
        \centering
        \includegraphics[height=0.63in,width=1.5in]{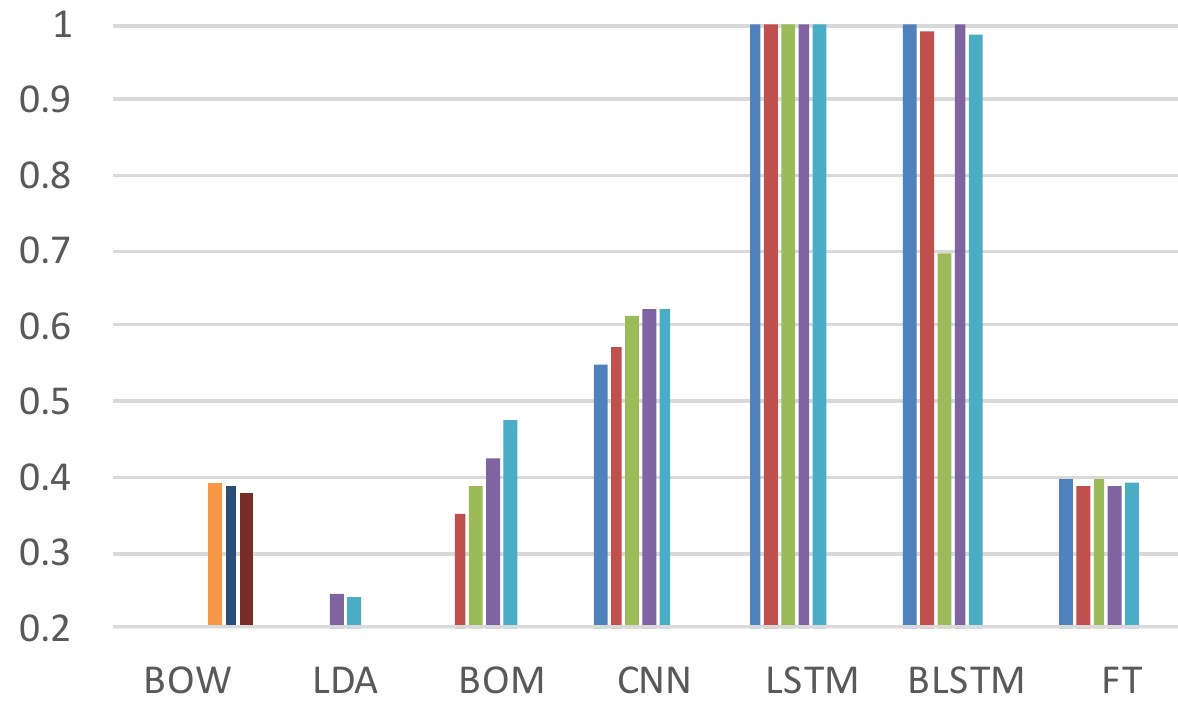}
        \caption{Length}
    \end{subfigure}
    ~ 
    \begin{subfigure}[t]{0.2\textwidth}
        \centering
        \includegraphics[height=0.63in,width=1.5in]{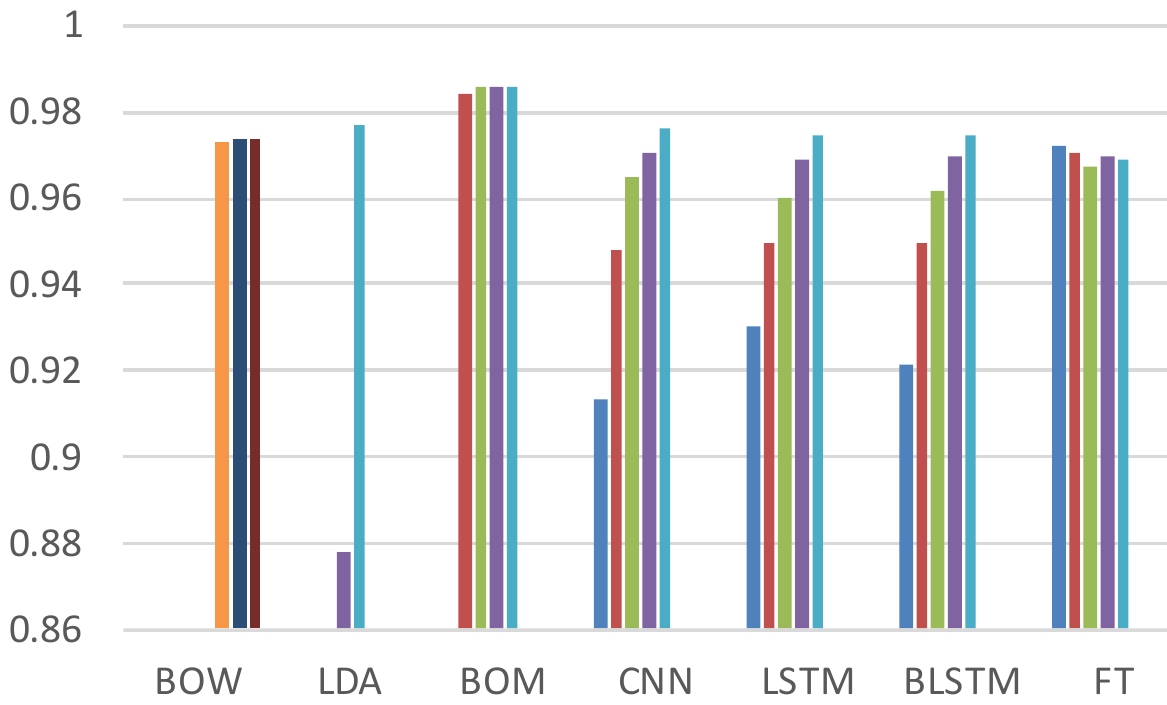}
        \caption{Content}
    \end{subfigure}
    ~
    \begin{subfigure}[t]{0.2\textwidth}
        \centering
        \includegraphics[height=0.63in,width=1.5in]{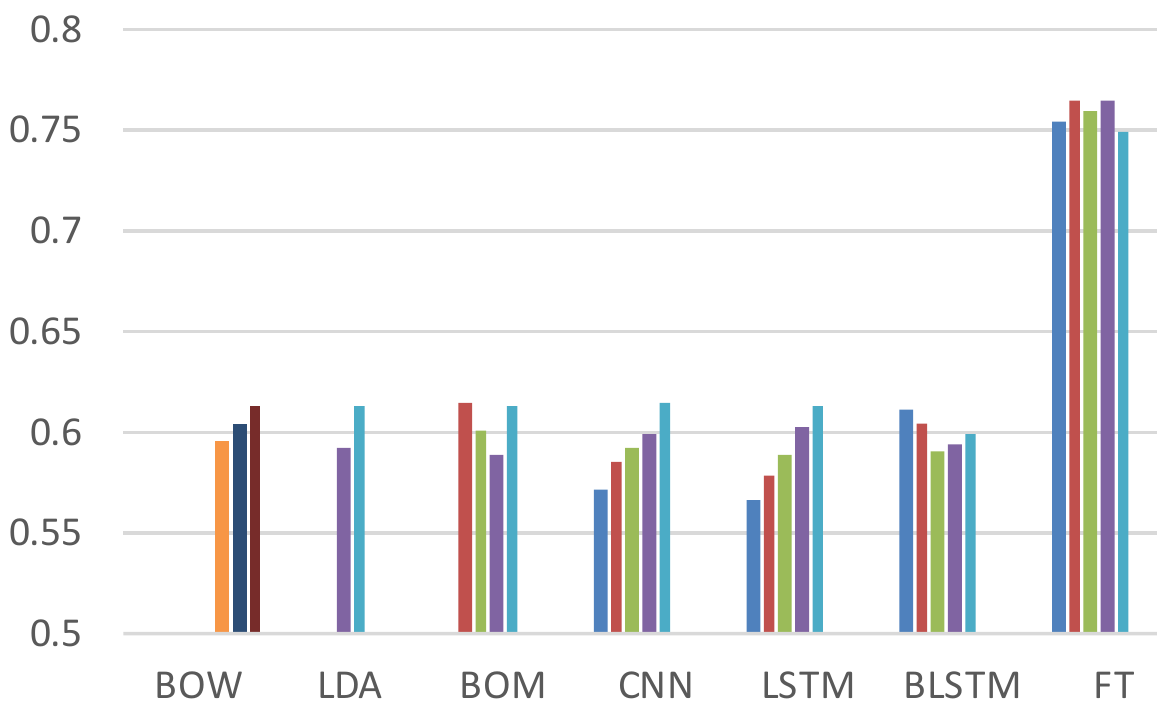}
        \caption{Word Order}
    \end{subfigure}
    ~
    \begin{subfigure}[t]{0.2\textwidth}
        \centering
        \includegraphics[height=0.63in,width=1.5in]{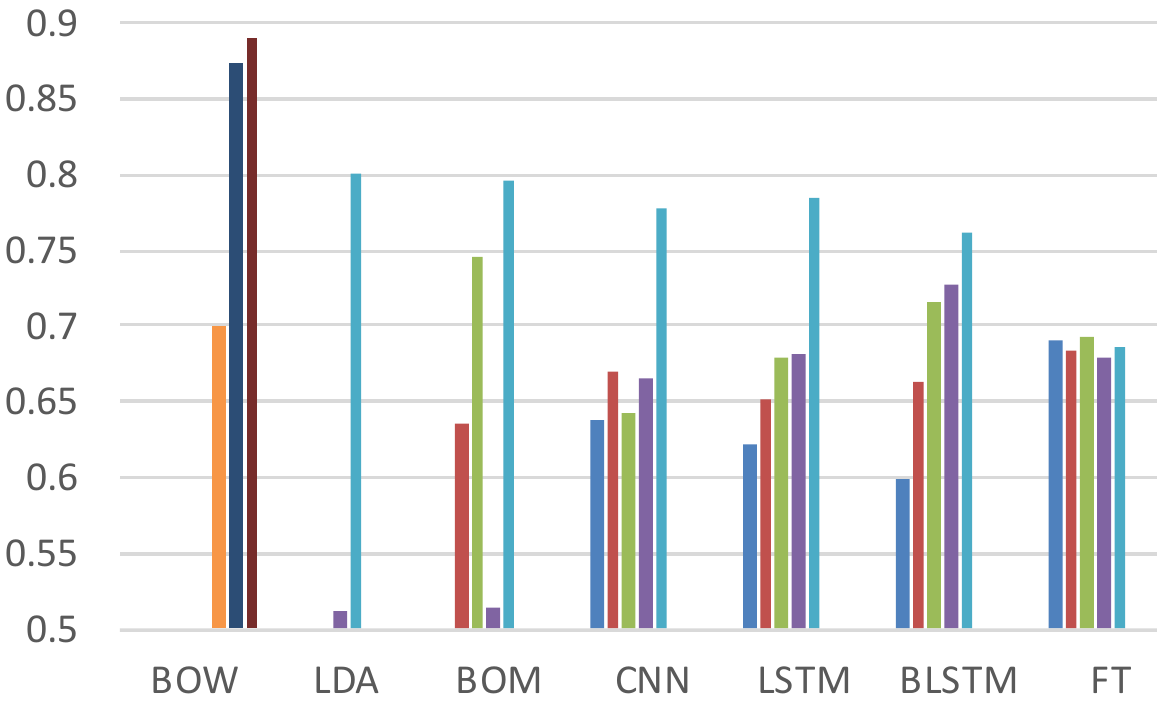}
        \caption{Slang Words}
    \end{subfigure}
    
    \begin{subfigure}[t]{0.2\textwidth}
        \centering
        \includegraphics[height=0.63in,width=1.5in]{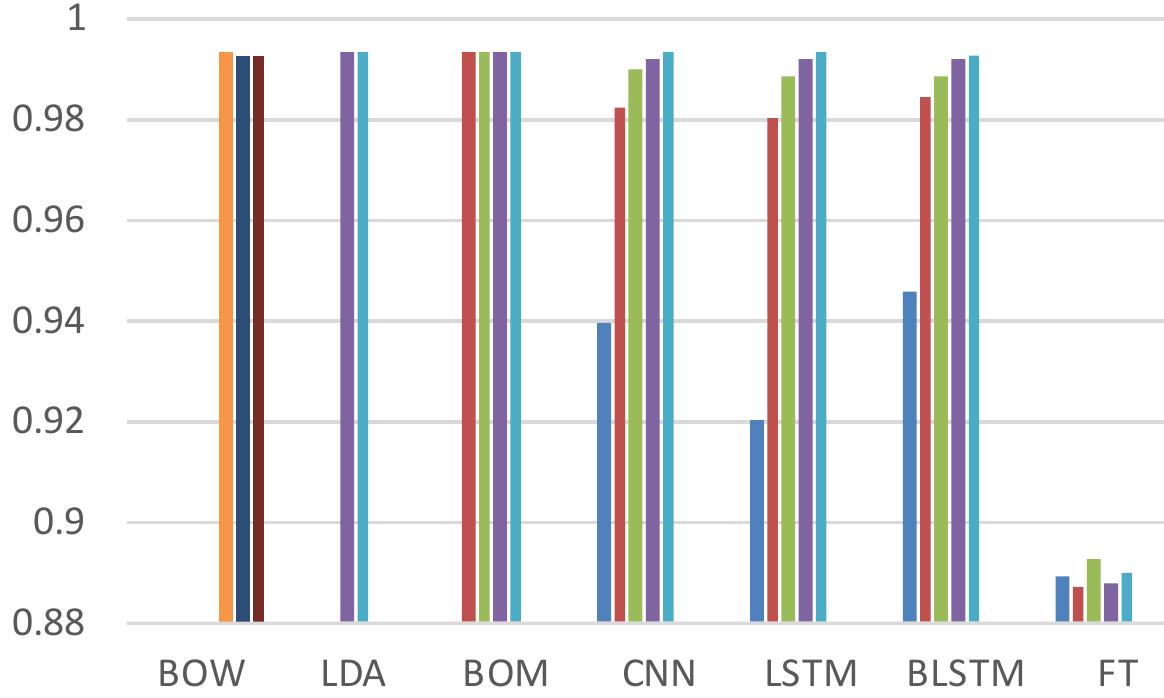}
        \caption{Hashtag}
    \end{subfigure}
    ~
    \begin{subfigure}[t]{0.2\textwidth}
        \centering
        \includegraphics[height=0.63in,width=1.5in]{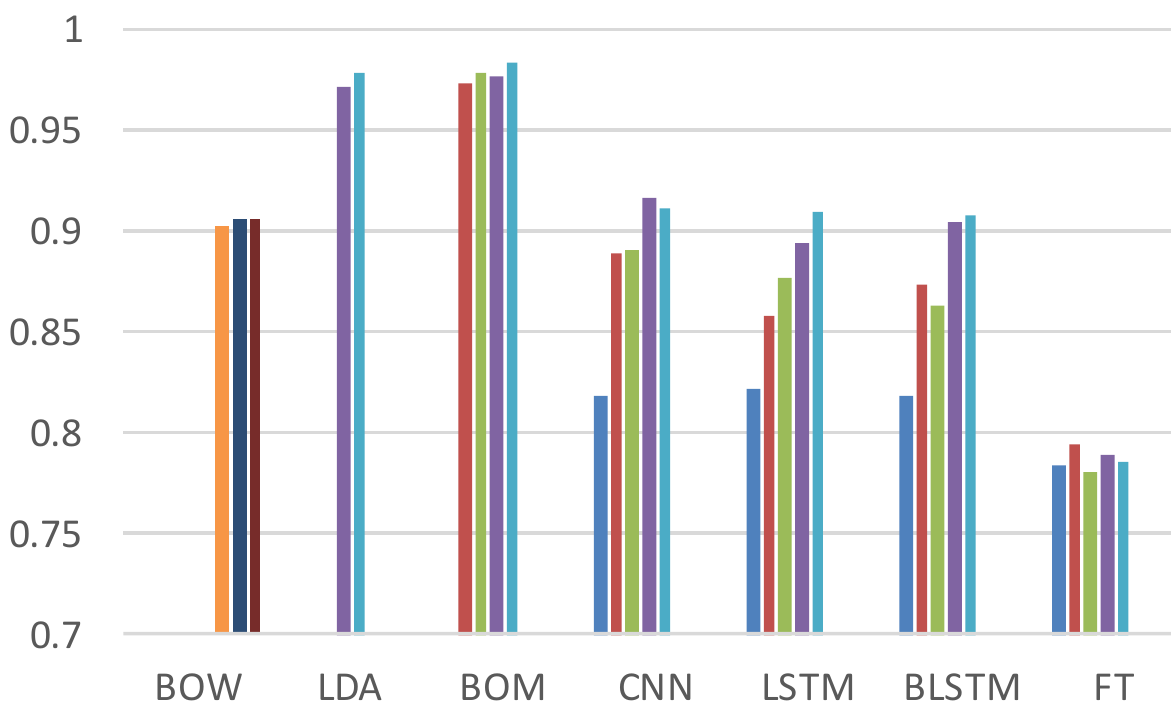}
        \caption{Named Entities}
    \end{subfigure}
    ~
    \begin{subfigure}[t]{0.2\textwidth}
        \centering
        \includegraphics[height=0.63in,width=1.5in]{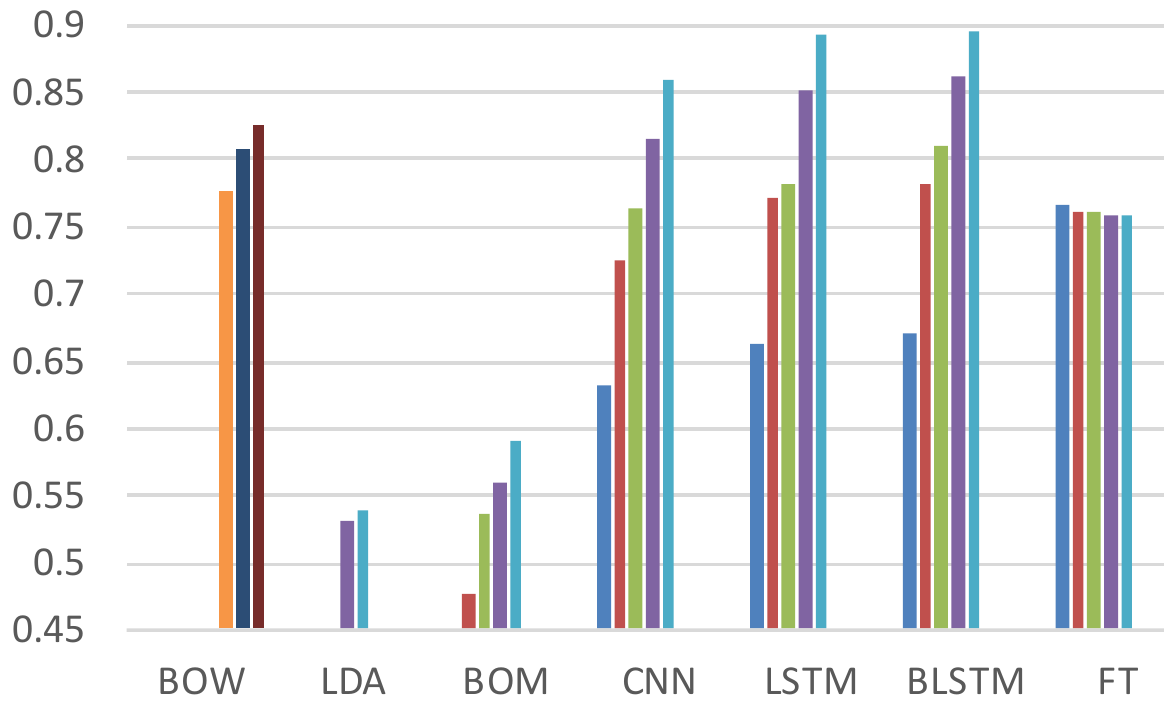}
        \caption{Capitalization Count}
    \end{subfigure}
    ~
    \begin{subfigure}[t]{0.2\textwidth}
        \centering
        \includegraphics[height=0.63in,width=1.5in]{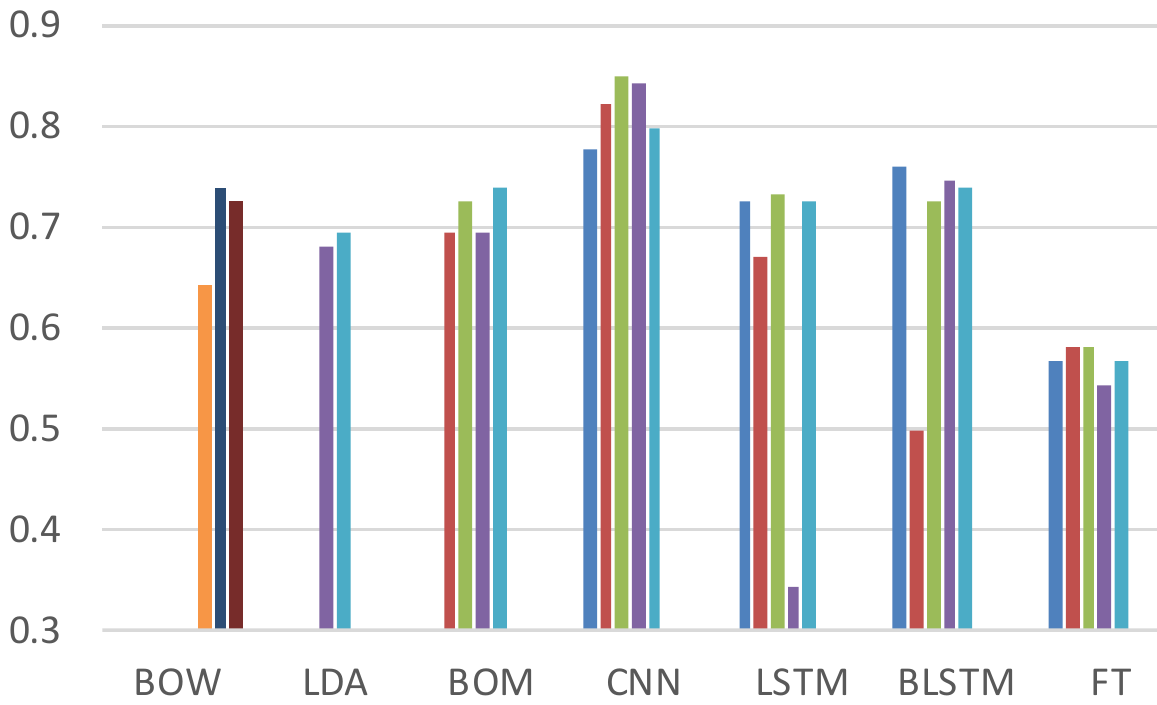}
        \caption{Informative Capitalization}
    \end{subfigure}    
    \begin{subfigure}[t]{0.2\textwidth}
        \centering
        \includegraphics[height=0.63in,width=1.5in]{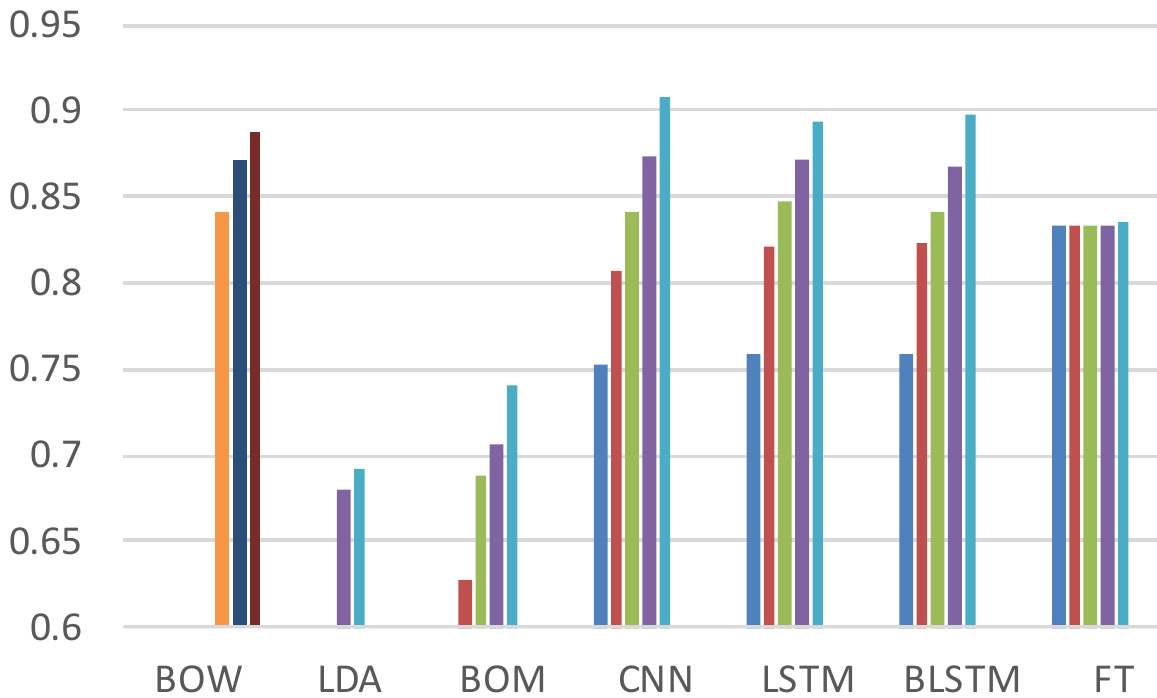}
        \caption{Mention Count}
    \end{subfigure}    
    ~
    \begin{subfigure}[t]{0.2\textwidth}
        \centering
        \includegraphics[height=0.63in,width=1.5in]{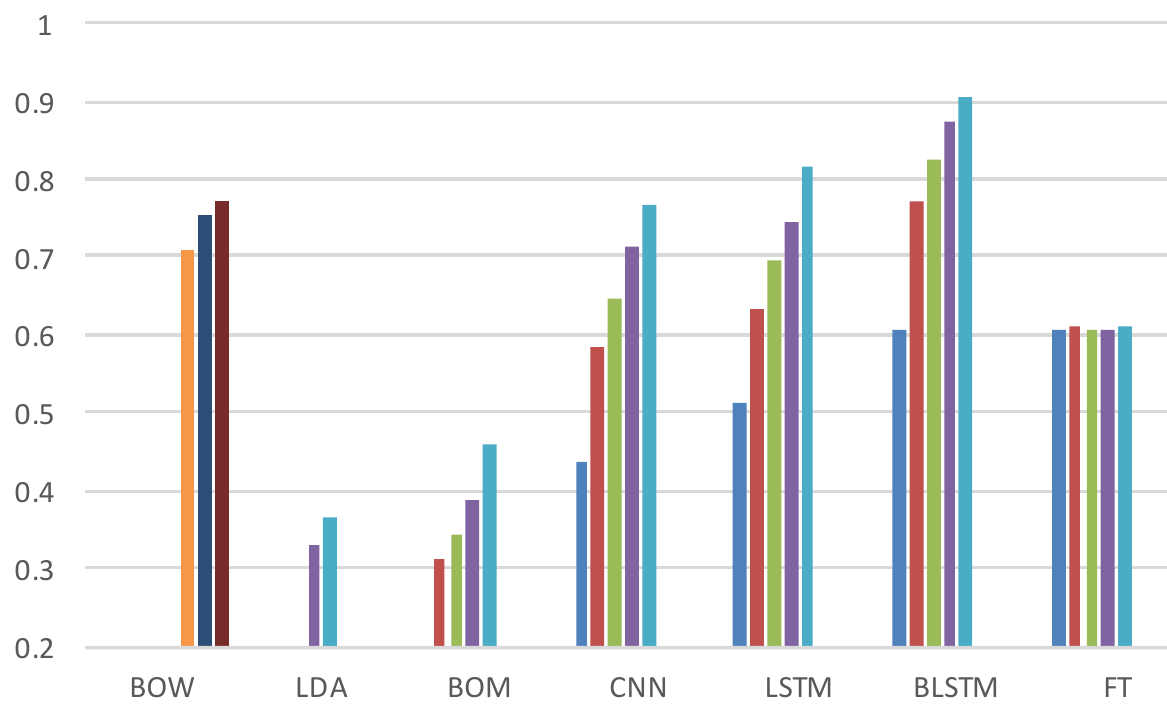}
        \caption{Mention Position}
    \end{subfigure}    
    ~
    \begin{subfigure}[t]{0.2\textwidth}
        \centering
        \includegraphics[height=0.63in,width=1.5in]{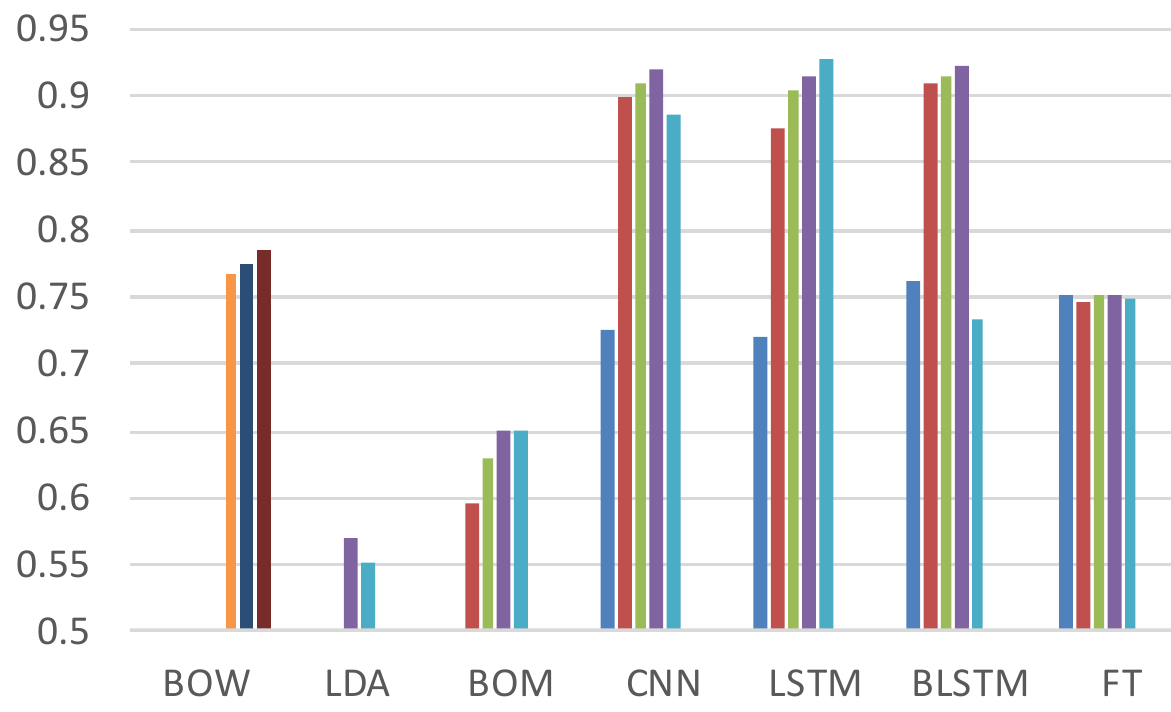}
        \caption{Is Reply}
    \end{subfigure}    
    ~
    \begin{subfigure}[t]{0.2\textwidth}
        \centering
        \includegraphics[height=0.63in,width=1.5in]{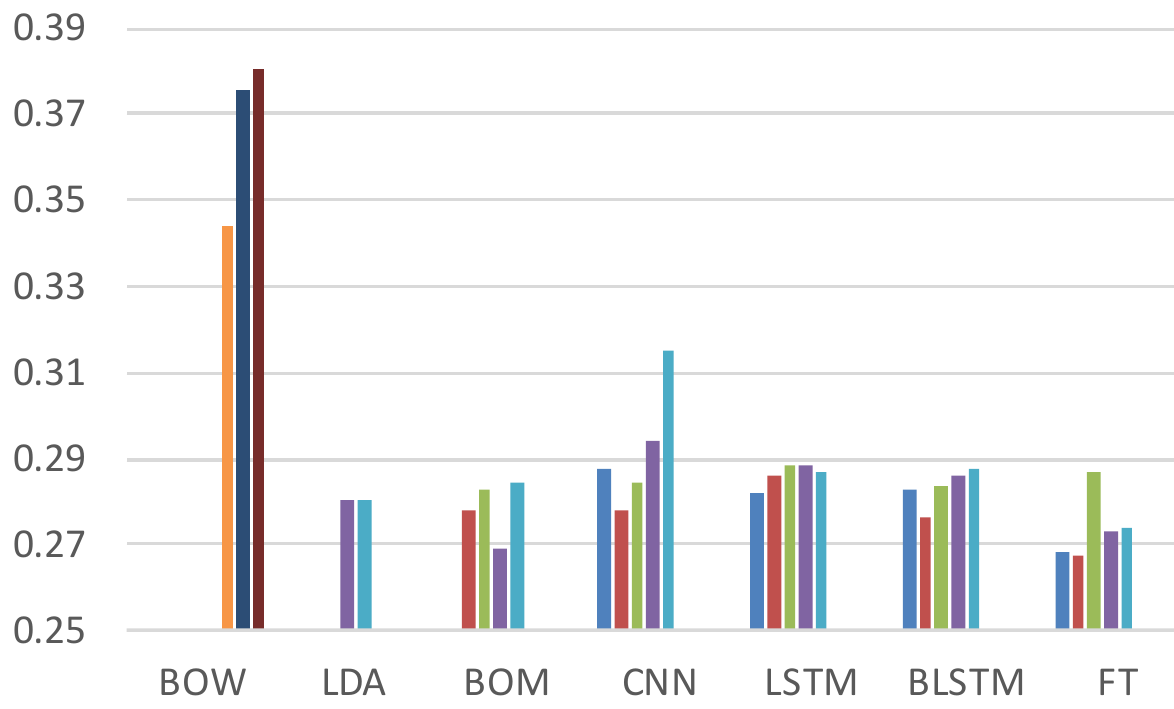}
        \caption{Reply Time}
    \end{subfigure}
    
    \begin{subfigure}[t]{0.4\textwidth}
        \centering
        \includegraphics[height=0.63in,width=2.5in]{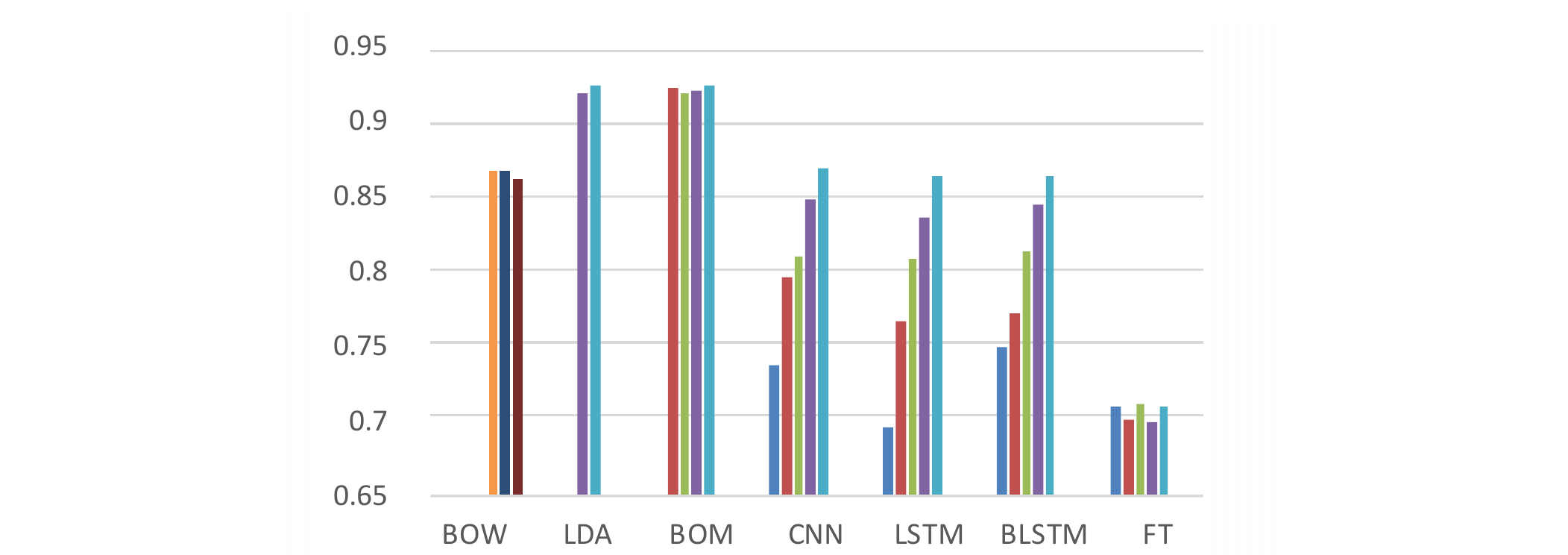}
        \caption{Word Repetition in Conversation}
    \end{subfigure}
    \begin{subfigure}[t]{1\textwidth}
        \centering
        \includegraphics[height=0.15in]{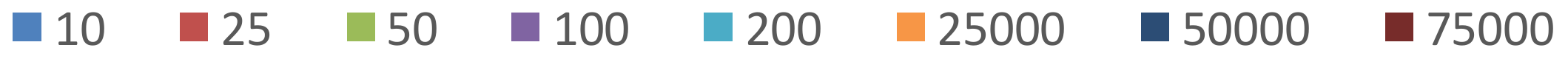}
    \end{subfigure}
    \caption{Sensitivity of the Models to Representation/Embedding Size}
    \label{fig:rs_p}
\end{figure*}

\subsection{Sensitivity to Representation Size}
In representation learning, low embedding size results in a poor model performance as the model does not have enough capacity (`underfits') to retain information. On the other hand, high embedding size also results in poor performance as the model has redundant bits of information (`overfits') which has a negative effect. The optimal strategy mostly is to do grid search for the size that gives superior performance. Specifically we build models with the embedding size from \{10, 25, 50, 100, 200\}. Figure~\ref{fig:rs_p} displays the plots for all the models for this setup. We find that performance of all the supervised models except FT is positively correlated with the representation size for most of the property prediction tasks. We discover that FT which relies on a simple operation of word vector average to represent a tweet is invariant to the representation size. This result is surprising as FT yields good performance with such a small embedding size of 10 (which indeed is the optimal hyper-parameter as suggested by the authors). We suggest to use FT for competitive performance on low-resource applications -- less memory (e.g., mobile), and faster computation. 


\subsection{Connections with the performance on downstream applications}
\label{a:conn_dt}
In this subsection, we will attempt to establish the correlation between the model performance on the various elementary property prediction tasks and the model performance on the various real applications.

\noindent\textbf{\underline{Sentiment Analysis}}: Giachanou et al.~\cite{giachanou16_csur} showed that sentiment analysis is typically aided by features such as content, slang words, mention count, hashtags and named entities. We observe that STV is the only unsupervised model to encode all the five task-specific relevant features well thereby outperforming the other models for this task. On the other hand, PV encodes relatively the least number of relevant features thereby faring poorer than BOW (as we see later) for this task. We find that most of the supervised models (excluding FastText) capture the task-specific features well.

\noindent\textbf{\underline{Hashtag Prediction}}: The salient features for hashtag prediction~\cite{tsur12_acl} include length, slang words and hashtag itself. We observe that none of the unsupervised models is able to encode all the relevant features. Since most of the supervised models are able to encode all the features, we conclude that nature of the task is strictly supervised.


\noindent\textbf{\underline{Named Entity Recognition}}: This task is benefitted by features such as slang words and capitalization~\cite{ritter11_emnlp}. This task also seems to be strictly supervised in nature as none of the unsupervised model is able to encode both the features. The recurrent models are able to capture both features successfully and clearly explains why these models are the state-of-the-art~\cite{zhang16_acl} for this task.

\noindent\textbf{\underline{Response Prediction}}: This is a social task to identify if a given tweet can receive a response~\cite{artzi12_naacl}. Content, hashtags and mention count are the vital features for this task. STV which is trained on a conversational context models all the features successfully. It is interesting to see that the recurrent models are also able to encode all the relevant features. This showcases the importance of recurrent models for the social tasks.

\begin{table}
\centering
\scriptsize
\caption{BOW versus Paragraph2Vec - Performance Comparison (F1-Score (\%)) on Multiple Tasks}
\label{tab:pv_vs_bow}
\begin{tabular}{|l|l|l|l|l|l|}
\hline
\textbf{Model/Task} & \textbf{SA} & \textbf{EI} & \textbf{TP} & \textbf{W} & \textbf{K} \\ \hline
BOW & 62.77 & 90.44 & 64.62 & 80.80 & 82.06 \\ \hline
Paragraph2Vec & 52.59 & 36.05 & 55.73 & 77.53 & 30.34 \\ \hline
\end{tabular}
\end{table}

\subsection{A Case Study of BOW vs Paragraph2Vec}
\label{a:bow_vs_p2v}
Table~\ref{tab:ta} shows that Paragraph2Vec is not good at encoding elementary tweet properties. To validate this with respect to high level applications, we compare Paragraph2Vec with BOW for a wide variety of Twitter applications. Specifically, we evaluate the models for five applications: (1) predict whether the sentiment of tweet is positive, negative or neutral (SA)~\cite{nakov16_semeval}, (2) predict the entity the tweet belongs to (EI)~\cite{amigo14_replab}, (3) predict the priority of the topic the tweet belongs to (TP)~\cite{amigo14_replab}, (4) predict the day of the weather referred in the tweet~\footnote{\url{https://www.kaggle.com/c/crowdflower-weather-twitter}} (W), and (5) predict the kind of the weather referred in the tweet (K). Table~\ref{tab:pv_vs_bow} reports the scores of the best performing Paragraph2Vec with the variant (BOW or Distributed Memory) and representation size (\{10, 25, 50, 100, 200\}) tuned using the validation set. From the results, we find that Paragraph2Vec  
has poor performance for all the tasks compared to BOW. Using this pair of models for various tasks, we have shown that performance of the models on the elementary tweet properties can help us estimate the performance of the models on various applications.

\subsection{Overall Insights}
Our extensive experimentation with a large number of models for important textual and social network properties of tweets, provides the following insights.
\begin{itemize}
\item Length prediction is the most difficult textual task while content prediction is the easiest. Word repetition is the easiest social task while reply time prediction is the most complicated.
\item Bi-directional LSTMs and Skip-Thought vectors (STV) best encode the textual and social properties of tweets respectively. Paragraph2Vec performs the worst. 
\item FastText is the best model for low resource applications providing very little degradation with reduction in embedding size. 
\item Relative performance of the models does not change based on tweet length. All models behave in the same way to variation in tweet length.
\end{itemize}

\section{Conclusion} 
\label{sec:conclusion}
In this paper, we tried to interpret multiple tweet representations in terms of the accuracy to which they encode elementary tweet properties (both textual and social). This helped us understand the weaknesses and strengths of such representations in an application independent, fine-grained manner. Based on such an evaluation, we conclude that Bi-directional LSTMs (BLSTMs) and Skip-Thought Vectors (STV) best encode the textual and social properties of tweets respectively. Also, FastText with huge information encoded in its small representation is the best model for low resource applications. In future, we plan to work on interpretation of distributed representations of nodes in a network wrt various interesting network properties.

\scriptsize
\bibliographystyle{IEEEtran}
\bibliography{biblioShort}

\end{document}